%% file: main.tex
\documentclass[10pt]{article}
\usepackage[accepted]{tmlr}
\usepackage{wrapfig}

\usepackage[pdfusetitle]{hyperref}
\usepackage{xcolor}
\hypersetup{
    colorlinks,
    linkcolor={red!50!black},
    citecolor={blue!50!black},
    urlcolor={blue!80!black}
}

\renewcommand{\headrulewidth}{0pt}
\usepackage{listings} 

\definecolor{llmgray}{gray}{0.9} 

\lstdefinestyle{llmoutput}{
    basicstyle=\ttfamily\small,
    frame=single,
    keepspaces=true,
  breaklines=true  
}

\usepackage[shortlabels]{enumitem}
\usepackage{microtype}
\usepackage{graphicx}
\usepackage{mdframed}
\usepackage{alltt}
\usepackage{booktabs} 
\usepackage{tablefootnote}
 
\usepackage{amsmath}
\usepackage{amssymb}
\usepackage{mathtools}
\usepackage{amsthm}
\usepackage{adjustbox}

\usepackage{tikz}
\usepackage{pgfplots}
\usepackage{subcaption}
\renewcommand{\paragraph}{\textbf}

\setlist{leftmargin=8mm}
\setlist[itemize]{leftmargin=5.5mm}
\theoremstyle{plain}

\theoremstyle{definition}

\theoremstyle{remark}

\usepackage[section]{placeins}
\title{Approaching Human-Level Forecasting with Language Models}

\author{\name Danny Halawi\texorpdfstring{$^*$}{} \hfill \texttt{\small dhalawi@berkeley.edu} \\
      \addr UC Berkeley
      \AND
      \name Fred Zhang\texorpdfstring{$^*$}{}\hfill \texttt{\small z0@eecs.berkeley.edu} \\
      \addr UC Berkeley
      \AND
      \name Chen Yueh-Han\texorpdfstring{$^*$}{}\hfill   \texorpdfstring{\texttt{\small john0922ucb@berkeley.edu}}{}\\
      \addr UC Berkeley
      \AND
      \name Jacob Steinhardt \hfill  \texorpdfstring{\texttt{\small jsteinhardt@berkeley.edu}}{}\\
            \addr UC Berkeley
      }

\begin{document}
\maketitle
\def\thefootnote{*}\footnotetext{Joint authorship.}\def\thefootnote{\arabic{footnote}}
  
\input{sec/abstract}
\input{sec/intro}
\input{sec/related_work}
\input{sec/preliminaries}
\input{sec/system}
\input{sec/optimization}

\input{sec/evaluations}
\input{sec/ablations}

\input{sec/conclusion}
\section*{Acknowledgments} We thank Jean-Stanislas Denain, Erik Jones, Ezra Karger, Jacob Pfau and Ruiqi Zhong for helpful discussions, and Jean-Stanislas Denain, Owain Evans, Dan Hendrycks, Horace He and Andy Zou for  comments and feedbacks on an early draft of the paper. DH was supported by an award from the C3.ai Digital Transformation Institute. FZ was supported by NSF award CCF-2311648.  JS was supported by the National Science Foundation SaTC CORE Award No. 1804794 and the Simons Foundation.

\bibliography{bib}
\bibliographystyle{apalike}

\newpage
\appendix
\input{appendix/models}

\input{appendix/base_eval}

\clearpage
\input{appendix/data}
\newpage
\input{appendix/system}
\clearpage
\input{appendix/opt}

\newpage
\input{appendix/examples}


\end{document}

%% file: sec/abstract.tex
\begin{abstract}
Forecasting future events is important for policy and decision making. In this work, we study whether language models (LMs) can forecast at the level of competitive human forecasters. Towards this goal, we develop a retrieval-augmented LM system designed to automatically search for relevant information, generate forecasts, and aggregate predictions. To facilitate our study, we collect a large dataset of questions from competitive forecasting platforms. Under a test set published after the knowledge cut-offs of our LMs, we evaluate the end-to-end performance of our system against the aggregates of human forecasts. On average, the system
nears the crowd aggregate of competitive forecasters, and in some settings surpasses it. 
Our work suggests that using LMs to forecast the future could provide accurate predictions at scale and help to inform institutional decision making.
\end{abstract}


%% file: sec/intro.tex
\section{Introduction}
\label{sec:intro}
Forecasting events is important in the modern world. Governments rely on  economic and geopolitical forecasts  for decision-making. 
Companies hire and  invest based on forecasts of  market conditions \citep{armstrong2001principles}.  In 2020, epidemiological forecasts for COVID-19 prompted national lockdowns across the globe \citep{adam2020special}.

There are two main approaches to forecasting. 
\textit{Statistical forecasting}  primarily uses tools from  time-series modeling. 
This methodology typically excels when data are abundant and    under minimal distributional shift. By contrast, in \textit{judgmental forecasting},   human forecasters assign probabilities to future events based on their own judgments, making use of historical data, domain knowledge, Fermi estimates, and intuition. They draw information from diverse sources and reason based on detailed contexts of the task. This enables accurate forecasts even with scarce past observations or under significant distributional shift \citep{tetlock2015superforecasting}.
We will refer to judgmental forecasting simply as ``forecasting''.

Since forecasting relies on human effort and expertise, it can be expensive, delayed, or applicable only in specific domains. Moreover, most human forecasts contain little or no explanatory reasoning. These limitations motivate  using language models (LMs) to automate forecasting \citep{hendrycks2021unsolved}. 
Because they can parse  
and produce texts rapidly, LMs can provide cheap and timely forecasts. Because they are pre-trained on web-scale data, they are endowed with massive, cross-domain knowledge. And  because we can elicit their reasonings through prompts, we can examine them to (partially) understand the final forecast.

In this work, we build a LM pipeline for automated forecasting, with a focus on predicting binary outcomes. Our system implements and automates three key components in  the traditional forecasting process: (1)   retrieval, which gathers relevant information from news sources; (2) reasoning, which weighs available data and makes a forecast; 
and (3) aggregation, which ensembles individual forecasts into an aggregated prediction. Each step makes use of an LM or a collection of LMs (either prompted or fine-tuned) (\autoref{fig:systems-overview}).

To optimize and evaluate our system, we collect a large dataset 
of forecasting questions from $5$ competitive forecasting platforms. The test set consists only of (binary) questions published after June 1st, 2023. Since this is after the knowledge cut-off date of our models, this prevents leakage from pre-training. The train set contains questions before June 1st, 2023, which we use for hyperparameter search and fine-tuning  our system.

We use a self-supervised approach to fine-tune a LM to make accurate predictions and explanatory reasonings. We first prompt a base LM with various scratchpads to elicit forecasts to questions in our training set. We then fine-tune a new LM on the outputs that outperformed the crowd, which teaches the model what reasoning method to apply in a given context and improves forecasting performance. For hyperparameter search, we identify system configurations, including retrieval and LM prompting strategies, that lead to the  best end-to-end performance.

\input{figures/retrieval_system}

Our optimized system approaches the performance of aggregated human forecasts over the test set, as measured by Brier score, a standard metric in forecasting. To our knowledge, this is the first automated system with forecasting capability that nears the human crowd level, which is generally stronger than individual human forecasters (\autoref{subsec:dataset}). We also consider a selective setting where our system uses heuristics, based on the LM's strengths, to decide whether to submit a forecast for a given question and date. 
In this setting, our system outperforms the human crowd. 

To summarize our main contributions:
\begin{enumerate}[itemsep=0pt,topsep=0pt,partopsep=0pt]
    \item We curate the largest, most recent dataset of real-world forecasting questions to date, for evaluating and optimizing automated forecasting systems. 
    \item We build a retrieval-augmented LM system that significantly improves upon the baseline and approaches the human crowd performance on competitive forecasting platforms.
    \item We propose and apply a self-supervised fine-tuning method to improve LM's capability in reasoning about forecasting tasks.
\end{enumerate}

%% file: figures/retrieval_system.tex
 \begin{figure*}[tbp]
\centering
\vspace{-1.1em}
\begin{subfigure}[b]{0.95\textwidth}
\includegraphics[width=\linewidth]{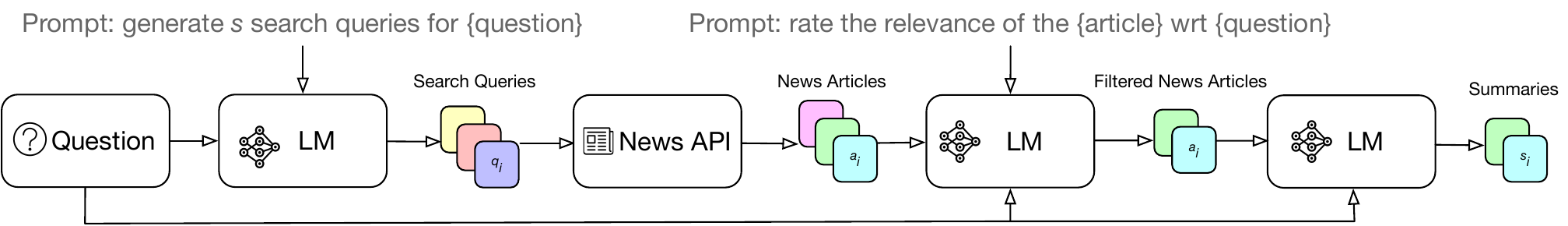}
\caption{\textbf{Our retrieval system}. The LM takes in the question and generates search queries to retrieve articles from historical news APIs. Then the LM ranks the articles on relevancy and summarizes the top $k$ articles.}
\label{fig:retrieval-system}
\end{subfigure}

\vspace{.5em}

\begin{subfigure}[b]{0.95\textwidth}
\includegraphics[width=\linewidth]{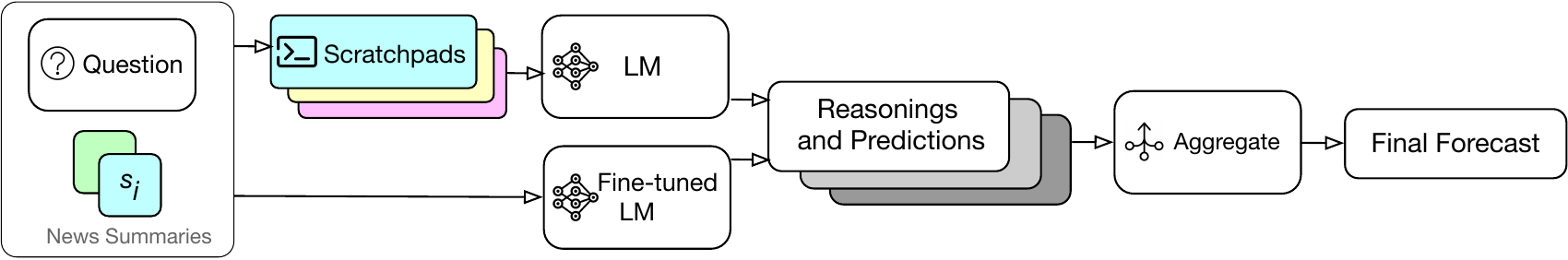}
\caption{\textbf{Our reasoning system}. The system takes in the question and summarized articles and prompts LMs to generate forecasts. The forecasts are then aggregated into a final forecast using the trimmed mean.}
\vspace{-0.2em}
\label{fig:reasoning-system}
\end{subfigure}
\caption{\textbf{Overview of our retrieval and reasoning systems}. Our retrieval system retrieves summarized new articles and feeds them into the reasoning system, which prompts LMs for reasonings and predictions that are aggregated into a final forecast.}
\label{fig:systems-overview}
\end{figure*}

%% file: sec/related_work.tex
\section{Related Work}
\label{sec:related_work}
\paragraph{Event forecasting.} Machine learning systems that make accurate, automated forecasts can help inform  human decision-making 
\citep{hendrycks2021unsolved}. 
\citet{jin2021forecastqa} provided ForecastQA, the first dataset for this task, which contains questions created by crowdworkers based on events from news articles. \citet{zou2022forecasting} introduced Autocast, a benchmark dataset compiled from forecasting competition questions up to 2022. In a competition with a large prize pool, no machine learning system was able to approach the performance of human forecasters on Autocast \citep{zou2022forecasting}. The knowledge cut-offs of LMs have moved past 2022, necessitating more recent data. In this work, we source questions in 2023--2024, enabling us to apply recent LMs.

\citet{yan2023autocast} built a retrieval system that led to improved accuracy on Autocast. They trained a  Fusion-in-Decoder  model to directly predict the final (binary) resolution \citep{izacard2021leveraging} and reported accuracy, whereas we elicit both explanatory reasonings and probability forecasts from LMs and measure performance with the standard Brier score metric.

\citet{schoenegger2023large,abolghasemi2023humans}
evaluated  GPT-4 and   other LLMs on forecasting tournaments and found that they underperform the human crowd.  This observation is in line with ours in \autoref{sec:0-shot}. Unlike us, they  make little or no efforts to improve these LMs on forecasting.

Finally, there has been recent work on using transformer models or LMs for statistical time-series forecasting \citep{nie2022time,gruver2023large,dooley2023forecastpfn,rasul2023lag,jin2023time,das2023decoder, woo2024unified}, but this is distinct from our focus on judgmental forecasting.

\paragraph{Information retrieval (IR).} IR can improve question-answering capabilities of LMs \citep{lewis2020retrieval, shuster2021retrieval, nakano2021webgpt}. In  event forecasting, access to diverse, up-to-date information is crucial \citep{tetlock2015superforecasting}. 
 Thus, a key component of our system is an IR architecture that furnishes the reasoning model with news articles, using LMs for query expansion, relevance ranking and summarization. Beyond our setting, using LMs for IR is an active research topic \citep{zhu2023large}.

\paragraph{Calibration.} Calibration is important for accurate forecasting \citep{tetlock2015superforecasting}. Hence, on competitive forecasting tournaments, forecasters are evaluated by proper scoring rules, such as Brier score \citep{brier1950verification}, which   incentivize calibration \citep{gneiting2007strictly}.  There is a vast literature on calibration in deep learning; see \citet{gawlikowski2021survey, wang2023calibration} for  surveys.

%% file: sec/preliminaries.tex
\section{Preliminaries: Data, Models and Baseline}
\label{sec:prelim}
\subsection{Dataset}
\label{subsec:dataset}
\paragraph{Data format.}
Forecasting platforms such as {Metaculus}, {Good Judgment Open}, {INFER}, {Polymarket}, and {Manifold} invite participants to predict future events by assigning probabilities to  outcomes of a question. 
Each question consists of a \textit{background description}, \textit{resolution criterion}, and  $3$ timestamps: a \textit{begin date} when the question was published, a  \textit{close date} when no further forecasts can be submitted, and (eventually) a \textit{resolve date} when the outcome is determined. 
A forecast can be submitted  between the begin date and min(resolve date, close date). See \autoref{tab: short-data-point} for an example question with these main fields.
\input{tables/short_sample_question}

\paragraph{Crowd prediction.}
On any given question, as individual forecasts are submitted, forecasting platforms continuously aggregate them into a crowd prediction; see \autoref{sec:crowd-pred} for details about the aggregation mechanisms.  The crowd prediction is a strong benchmark to compete with. For example, \citet{Metaculus_2023} shows that an ensemble of all forecasters consistently outperforms using just the top 5, 10, ..., 30 best forecasters (based on past scores). In this work, we compare our system performance to the crowd aggregates. 

\paragraph{Raw data.}
We source forecasting questions from the 5 above-mentioned platforms. This yields a total of $48{,}754$ questions and $7{,}174{,}607$ user forecasts spanning from 2015 to 2024. The dataset includes $33{,}664$ binary questions, $9{,}725$ multiple-choice questions, $4{,}019$ numerical questions, and $1{,}346$ questions of other types. 
The questions cover a wide range of topics across the globe (\autoref{fig:location-mentions}).

The raw dataset contains questions that are ill-defined, overly personal, or of niche interests. 
Furthermore,  recent questions are highly unbalanced, with over $80\%$ of questions since June 1, 2023 coming from Manifold and Polymarket.

\paragraph{Data curation.}
To address the above issues, we curate a subset by filtering ill-defined questions and removing questions that received few forecasts or trading volume on Manifold and Polymarket. We focus on predicting binary questions and split multiple-choice questions into binary ones.

To guard potential leakage from LMs' pre-training, we only include questions in the test set that appear after the knowledge cut-off for the models we use (June 1, 2024). 
All test set questions were opened after the date, and  all train and validation questions were resolved before. Questions that span across the date are discarded. 

This yields a set of $5{,}516$ binary questions, including $3{,}762$ for training, $840$ for validation, and $914$ for testing (\autoref{tab:curated-split}). See \autoref{tab: sample-data-point} for a sample data point and \autoref{sec:data-details} for details about the   curation process.

\input{figures/2-table}
\subsection{Evaluation}
\paragraph{Retrieval schedule.} 
We can simulate forecasting the future by leveraging the fact that models are only trained up to a cut-off date \citep{zou2022forecasting}. To simulate a forecast for a question that has been resolved, we query a historical news corpus to retrieve articles between the question begin date and a specified \textit{retrieval date} \citep{zou2022forecasting,yan2023autocast}. The retrieval date can be viewed as the “simulated date” of the forecast, as we are mimicking the information the model would have had access to on that date. 

To create a set of retrieval dates for each question, we use geometrically increasing time points between the open and close dates. We choose this schedule for two reasons: (1) questions are often most active shortly after they open, and (2) some questions have overly conservative close dates that are long after the question resolves. We use $n=5$ retrieval dates per question; the $k$th  retrieval date 
is calculated as  
\begin{equation}%
\label{eqn:schedule}
\text{retrieval\_date}_k = \text{date}_\text{begin} +  {(\text{date}_\text{close} - \text{date}_\text{begin} - 1)^{k/n}}.
\end{equation}
For questions that resolve before they close, we exclude all dates occurring after the question has been resolved. Under this geometric retrieval schedule, we retain $86\%$ of retrieval dates  on average across all questions (\autoref{fig:retrieval-retention}).
The average question window in our corpus is approximately $70$ days, and the average time until resolution is $42$ days.

In our dataset,  questions can get resolved long before their official close date. This occurs for questions like ``Will $\langle event \rangle$ happen by $\langle date\rangle$'', where resolving early indicates that  the event did occur (see \autoref{tab: short-data-point} for an example). 
It is tempting to choose retrieval dates with respect to the resolve date so that each question can receive the same number of retrieval dates, e.g.~by retrieving at geometric intervals between the open and resolve date. However, this would leak information, since the retrieval date would now depend on the resolve date, which, as we explained, correlates with the resolution.  

\paragraph{Metric.}
Our work focuses on binary questions and uses the Brier score as the performance metric, defined as $(f-o)^2$, where $f\in [0,1]$ is the probabilistic forecast and $o\in\{0,1\}$ is the outcome. The Brier score is a strictly proper scoring rule: assuming the true probability that $o=1$ is $p$, the optimal strategy is to report $f=p$. This is desirable, since improper scoring rules would incentivize reporting distorted probabilities. As a baseline, an (unskilled) forecast of $.5$ attains a Brier score of $.25$. 

To compute the final Brier score, we first average the Brier scores across retrieval dates for each question, then average across questions. 
%
%
We also report standard errors; however, note that the computation of standard errors assumes the data are i.i.d., while our data are in fact time-series, so this likely underestimates the true error. Finally, we also measure calibration with root mean square (RMS) calibration error.
  
\subsection{Models}
We evaluate $14$ instruction-tuned LMs: GPT-3.5-Turbo, GPT-3.5-Turbo-1106 \citep{brown2020language}; GPT-4, GPT-4-1106-Preview \citep{achiam2023gpt}; Llama-2-7B, Llama-2-13B, Llama-2-70B \citep{touvron2023Llama}; Mistral-7B-Instruct, Mistral-8x7B-Instruct \citep{jiang2024mixtral}, Nous Hermes 2 Mixtral-8x7B-DPO, Yi-34B-Chat, Claude-2, Claude-2.1 \citep{Claude}, and Gemini-Pro \citep{team2023gemini}; see \autoref{sec:appendix-models} for details. 

\subsection{Models are not naturally good at forecasting}
\label{sec:0-shot}
As a baseline, we evaluate all $14$ LMs  with no additional information retrieval. We use zero-shot prompts and scratchpad prompts \citep{nye2021show}. For each prompting strategy, we craft candidate prompts, pick the best prompt on the validation set, and report its Brier scores on the test set. The   results are given in  \autoref{tab:model_comparison_short}, where we report the best model in each series; see \autoref{tab:model_comparison} for full statistics.
The  prompt choices appear  in \autoref{fig:zero-shot-base-eval-prompt} and \autoref{fig:scratchpad-base-eval-prompt} and further details are in \autoref{sec:base-eval}.

None of the models are naturally good at forecasting. Most models' scores are around or worse than random guessing ($.25$). Only the GPT-4 and Claude-2 series beat the unskilled baseline by a large margin ($>.02$). 
Moreover, while  GPT-4-1106-Preview achieves the lowest Brier score of $.208$, it  trails significantly behind the human crowd performance of $.149$. 

%% file: tables/short_sample_question.tex
\begin{table*}[t!]
  \centering
  \small
  \renewcommand{\arraystretch}{1.3}
  \setlength\tabcolsep{4pt}
  \begin{tabular}{lp{0.75\textwidth}}
      \toprule
      \textbf{Field} & \textbf{Content} \\
      \midrule
      Question & Will Starship achieve liftoff before Monday, May 1st, 2023? \\
      Background & On April 14th, SpaceX received a launch license for its Starship spacecraft. A launch scheduled for April 17th was scrubbed due to a frozen valve. SpaceX CEO Elon Musk tweeted: ``Learned a lot today, now offloading propellant, retrying in a few days \ldots'' \\
      Resolution Criteria & This question resolves Yes if Starship leaves the launchpad intact and under its own power before 11:59pm ET on Sunday, April 30th. \\
   Key Dates & Begin Date: 2023-04-17  \quad $\mid$ \quad   Close Date:  2023-04-30 \quad $\mid$ \quad Resolve Date: 2023-04-20 \\
      \bottomrule
  \end{tabular}
  \caption{\textbf{A sample question} with its background, resolution criteria, and key dates. The question resolved early (with a final resolution of Yes). See \autoref{tab: sample-data-point} for the complete sample point.} 
  \vspace{-1.1em}
  \label{tab: short-data-point}
\end{table*}

%% file: figures/2-table.tex
\begin{table}[!t]
    \centering
    \begin{subtable}[b]{0.44\textwidth}
   \centering
  \begin{tabular}{lccc}
    \toprule
    \textbf{Platform} &  \textbf{Train} & \textbf{Validation} & \textbf{Test} \\
    \midrule
    Metaculus  & $1,576$ & $230$ & $275$ \\
    GJOpen  & $806$ & $161$ & $38$ \\
    INFER & $52$ & $50$ & $4$ \\
    Polymarket & $70$ & $229$ & $300$ \\
    Manifold   & $1,258$ & $170$ & $297$ \\
    \midrule
    \textbf{All Platforms} & $\mathbf{3,762}$ & $\mathbf{840}$ & $\mathbf{914}$ \\
    \bottomrule
  \end{tabular}
  \caption{Dataset distribution}
  \label{tab:curated-split}
 \end{subtable}%
 \hspace{0.45em}%
    \begin{subtable}[b]{0.55\textwidth}
    \centering
\begin{tabular}{@{}l r r@{}}
    \toprule
    \textbf{Model} & \textbf{Zero-shot} & \textbf{Scratchpad} \\
    \midrule
    GPT-4-1106-Preview & $\mathbf{0.208}$ $(0.006)$ & $\mathbf{0.209}$ $(0.006)$ \\
    Llama-2-13B & $0.226$ $(0.004)$ & $0.268$ $(0.004)$ \\
    Mistral-8x7B-Instruct & $0.238$ $(0.009)$ & $0.238$ $(0.005)$ \\
    Claude-2.1 & $0.220$ $(0.006)$ & ${0.215}$ $(0.007)$ \\
    Gemini-Pro & $0.243$ $(0.009)$ & ${0.230}$ $(0.003)$ \\
    \midrule
    Trimmed mean & $0.208$ $(0.006)$ & $0.224$ $(0.003)$ \\
    \bottomrule
  \end{tabular}
  \caption{Baseline performance of pre-trained models}
  \label{tab:model_comparison_short} 
    \end{subtable}
    \caption{\textbf{(a)} \textbf{Distribution of our train, validation, and test sets across all 5 forecasting platforms.} Importantly, every question in the test set is from June 1, 2023 or later, after the training cut-off of our base LMs. Meanwhile, all questions in the train and validation sets were resolved before June 1, 2023, ensuring   no leakage from the tuning process.  \textbf{(b)} \textbf{Baseline performance} of pre-trained models on the test set, with $1$ standard error (SE) (see full results in \autoref{tab:model_comparison}). Random baseline: 0.250; human crowd: 0.149. The results underscore that models are not naturally good at forecasting.}
    \label{fig:2-table}
\end{table}

%% file: sec/system.tex
\section{Our System}
\label{sec:system}

\input{tables/eval_selective_setting}
As observed in \autoref{tab:model_comparison_short}, all models perform poorly in the baseline setting.
We intuit that  models require detailed contexts and up-to-date information to make accurate forecasts. Our system addresses this issue via news retrieval and elicits better reasoning via optimized prompting strategies and fine-tuning.

\subsection{Retrieval}
\label{sec:retrieval-sys}
Our retrieval system consists of $4$ steps: search query generation, news retrieval, relevance filtering and re-ranking, and text summarization (\autoref{fig:retrieval-system}).

First, we generate search queries that are used to invoke news APIs to retrieve historical articles. We initially implement a straightforward query expansion prompt (\autoref{fig:sq-prompt-2}), instructing the model to create queries based on the question and its background. However, we find that this overlooks sub-considerations that often contribute to accurate forecasting. To achieve broader coverage, we prompt the model to decompose the forecasting question into sub-questions and use each  to generate a search query  \citep{min2019multi};  see \autoref{fig:sq_prompt} for the prompt.
For instance, when forecasting election outcomes, the first approach searches directly for polling data, while the latter creates sub-questions that cover campaign finances, economic indicators, and geopolitical events. We combine both approaches for comprehensive coverage.

Next, the system retrieves articles from news APIs using the LM-generated search queries. 
We evaluate $5$ APIs on the relevance of the articles retrieved and select NewsCatcher\footnote{\href{https://www.newscatcherapi.com/}{https://www.newscatcherapi.com/}} and Google News (\autoref{sec:news-api-eval}).

Our initial retrieval provides wide coverage at the cost of obtaining some irrelevant articles. To ensure that they do not mislead the model at the reasoning step, we   prompt GPT-3.5-Turbo to rate the relevancy of all articles (\autoref{fig:relevance-prompt}) and filter out low-scoring ones. Since the procedure is costly in run-time and budget, we only present the article's title and first 250 words to the model in context. We validate that this approach achieves high recall and precision while saving $70\%$ cost (see \autoref{sec:relevance} for   alternative methods  and   results). 

Since LMs are limited by their context window, we summarize the articles. In particular, we prompt     GPT-3.5-Turbo to distill the most relevant details from each article with respect to the forecasting question (\autoref{fig:summarization-prompt}). Finally, we present the top $k$ article summaries to the LM, ordered by their relevancy. We choose the ranking criterion, article count $k$, and summarization prompt based on end-to-end Brier scores over the validation set; see \autoref{subsec:hp_sweep} for the hyperparameter sweep procedure.

\subsection{Reasoning}
Prior work in forecasting has focused on eliciting predictions from models without requiring rationales \citep{zou2022forecasting, yan2023autocast}. However, accurately predicting the future is a difficult task that often requires computation beyond a single forward pass. Having the model externalize its reasoning also allows us to understand the explanation for the forecast and improve it accordingly.

We use open-ended scratchpad to structure model's reasoning paths. Our prompt begins with posing the question, providing a description,  and specifying resolution criteria and key dates, followed by the top $k$ relevant summaries (\autoref{fig:basic-prompt-reasoning}).  To guide the model to reason about the forecasting question, the optimal scratchpad prompt (\autoref{fig:optimal-reasoning-prompt}), as identified in \autoref{subsec:hp_sweep}, also incorporates four additional components:
\begin{itemize}
\item 
First, to ensure that the model comprehends the question, we prompt it to rephrase the question. It is also instructed to expand the question with its own knowledge to provide further information.  Intuitively, a more detailed and precise phrasing of the question elicits better responses \citep{deng2023rephrase}.
\item 
Forecasting requires a holistic consideration of the possibilities \citep{tetlock2015superforecasting}. We next prompt the model to leverage the retrieved information and its pre-training knowledge to produce arguments for why the outcome may or may not occur. 
\item 
The model can potentially generate weak arguments. To avoid treating all considerations as equal, it is instructed to weigh them by importance and aggregate them accordingly into an initial forecast. 
\item 
Finally, to prevent potential bias and   miscalibration, the model is asked to check if it is over- or under-confident and consider historical base rates \citep{tetlock2015superforecasting}, prompting it to calibrate and amend the prediction accordingly.
\end{itemize}

\textbf{Base model.} We prompt GPT-4-1106-Preview with the best scratchpads (found via hyperparameter sweep), since it consistently gives the lowest Brier scores among the LMs we test (see \autoref{subsec:hp_sweep} on reasoning).

\textbf{Fine-tuned model.} We also prompt a fine-tuned version of GPT-4 that we trained to generate reasonings with accurate predictions (\autoref{subsec:fine_tuning}). We prompt it with only the question's basic information (no scratchpad instructions) since our fine-tuned model is trained to reason without prescriptive instructions.

\subsection{Ensembling}
Since the aggregate of predictions is usually superior to individual forecasts \citep{tetlock2015superforecasting}, we elicit multiple predictions from the base and fine-tuned models.

We prompt GPT-4-1106-Preview with the optimal scratchpad prompt (\autoref{fig:optimal-reasoning-prompt}), along with the $2$ next best scratchpad prompts identified in \autoref{subsec:hp_sweep}. For our fine-tuned model, we set temperature $T = 0.5$ and prompt it $3$ times to sample $3$ additional forecasts. This gives us $6$ forecasts in total: $3$ from the base model, and $3$ from the fine-tuned model.
Given these forecasts, the system ensembles them into a final prediction by taking their trimmed mean, as this performs best on the validation set among the ensemble methods we implement (see \autoref{subsec:hp_sweep} on ensembling).

We provide further details about our system in  \autoref{sec:system-details}, including hyperparameters and prompt designs.

%% file: tables/eval_selective_setting.tex
\begin{table}[!t]
    \centering
    \resizebox{\textwidth}{!}{%
    \begin{tabular}{lcccccccc}
      \toprule
      \textbf{Criteria} & \multicolumn{3}{c}{\textbf{Brier Score} $\downarrow$} & \multicolumn{3}{c}{\textbf{\% Accuracy} $\uparrow$} & \multicolumn{2}{c}{\textbf{\% Data Retained} $\uparrow$} \\
      \cmidrule(lr){2-4} \cmidrule(lr){5-7}  \cmidrule(lr){8-9}
      & \textbf{Ours} & \textbf{Crowd} & \textbf{Aggregate} & \textbf{Ours} & \textbf{Crowd} & \textbf{Aggregate} & \textbf{Forecasts} & \textbf{Questions} \\
      \midrule
      \textbf{All Questions} & $.179_{.003}$ & $.149_{.003}$ & $\underline{\mathbf{.146_{.002}}}$ & $71.5_{.7}$ & $77.0_{.7}$ & $\underline{\mathbf{77.8_{.6}}}$ & $100\%$ & $100\%$ \\
      \textbf{Crowd Uncertain} & $\mathbf{.238_{.004}}$ & $.240_{.003}$ & $\underline{\mathbf{.233_{.002}}}$ & $58.1_{1.3}$ & $58.3_{1.3}$ & $\underline{\mathbf{60.2_{1.2}}}$ & 51\% & 56\% \\
      \textbf{Early Retrieval} & $.186_{.003}$ & $.162_{.004}$ & $\underline{\mathbf{.159_{.003}}}$ & $70.0_{.9}$ & $74.4_{.9}$ & $\underline{\mathbf{75.0_{.8}}}$ & 84\% & 100\% \\
      \textbf{$\mathbf{5+}$ Articles} & $.175_{.003}$ & $.142_{.003}$ & $\underline{\mathbf{.140_{.002}}}$ & $72.3_{.8}$ & $77.7_{.7}$ & $\underline{\mathbf{78.7_{.7}}}$ & 84\% & 94\% \\
      \midrule
      \textbf{All Criteria} & $\mathbf{.240_{.005}}$ & $.247_{.004}$ & $\underline{\mathbf{.237_{.003}}}$ & $\underline{\mathbf{58.0_{1.7}}}$ & $54.2_{1.7}$ & $\mathbf{56.6_{1.7}}$ & 22\% & 43\% \\
      \bottomrule
      
    \end{tabular}
    }
    \caption{\textbf{System performance} on the test set. ``All Questions'' shows the Brier score on the full test set. Other rows show selective evaluation when specified criteria are met, averaging over qualifying questions and retrieval dates. ``Crowd Uncertain'' refers to questions with crowd predictions between 0.3-0.7. ``Early Retrieval'' refers to the first 3 retrieval dates. ``5+ Articles'' refers to forecasting when at least 5 relevant articles are retrieved. Finally, ``All Criteria'' refers to forecasting when the $3$ criteria are jointly met. Notably, in every setting the aggregate (average) of our system and crowd prediction is the best. Subscript numbers indicate $1$ standard error. We bold entries that outperform the crowd aggregate, and underline the best entry in each category.}
    \label{tab:eval-selective-setting}
\end{table}

%% file: sec/optimization.tex
\input{figures/fine_tuning_diagram}
\section{Optimizing the System}
\label{sec:optimization}
We now describe the procedure to optimize our retrieval and reasoning system and the results obtained.

\subsection{Fine-tuning a Reasoning Model}
\label{subsec:fine_tuning}
We fine-tune a LM to produce reasonings that lead to accurate forecasts. To generate the data for fine-tuning, we (1) collect a large set of forecasts on the train set, and then (2) select a subset where the model outperforms the human crowd.

\paragraph{Collecting fine-tuning data.} To generate the preliminary data, we run our system at each retrieval date in the retrieval schedule and on each question in the train set, multiplied by 16 configurations described below.

First, as a form of data augmentation, we retrieve 2 sets of articles for each question by sampling 2 (distinct) retrieval configurations (\autoref{fig:fine-tuning-diagram}, left). Specifically, we sample the retrieval prompt, number of queries, and articles per query, twice (\autoref{sec:system}), with relevancy filtering and summarization following the process described in \autoref{sec:retrieval-sys}. This results in 2 inputs to the reasoning model per question, each with the same question but a different set of articles. 

To increase the chance of attaining a prediction that outperforms the crowd, we generate 4 candidate outputs per input (8 total per question) by trying different scratchpad prompts. The first uses the optimal prompt found in \autoref{subsec:hp_sweep} (\autoref{fig:optimal-reasoning-prompt}). We then sample 3 other scratchpad prompts, with probability inversely proportional to their Brier score on the validation set. We prompt both Claude-2.1 and GPT-4-Preview, since we find that Claude-2.1 is better on some questions. In total, this gives 16 candidate forecasts  per question.

\paragraph{Selecting fine-tuning data.}
We seek to fine-tune our model on strong forecasts.
To select the  data, we only keep outputs that give a lower Brier score than the crowd's. However, this can inadvertently cause overconfidence in our fine-tuned model. To mitigate this, we discard pairs where the prediction deviates by more than $0.15$ from the crowd prediction, and we also average our prediction with the crowd prediction when constructing the target output.

The resulting fine-tuning data has the following structure (\autoref{fig:fine-tuning-diagram}, right):
\begin{itemize}
    \item The \textbf{input} to the model consists of the question, description, and resolution criteria, followed by summarized articles.
    \item The \textbf{target output} consists of a reasoning and a prediction.
\end{itemize}

Importantly, the fine-tuning input excludes the scratchpad instructions. By doing so, we directly teach the model which reasoning  to apply in a given context. 

In total, $73{,}632$ reasonings are generated from which $13{,}253$ meet the above desiderata. Finally, we fine-tune GPT-4-0613\footnote{While the more recent GPT-4-1106-Preview has 2 years of more recent knowledge, it was not available for fine-tuning.} on the $6{,}000$ most recent points for 2 epochs, due to budget constraint
(\autoref{fig:fine-tuning-diagram}, right).

\subsection{Hyperparameter Sweep}
\label{subsec:hp_sweep}
Our hyperparameter sweep optimizes an (intermediate) metric over a discrete set of choices, such as prompts and the number of articles presented. We share the key findings below and  more details in \autoref{sec:appendix-opt}.

\paragraph{Methodology.}
We divide the hyperparameters into groups of 1-2 and optimize them iteratively. For each group, we select the best configuration based on the average Brier score on the validation set, except for search query generation where we use proxy metrics for efficiency.

We optimize the groups sequentially, fixing the optimal configurations from previous groups while sweeping the current one. The hyperparameters yet to be swept are randomized independently for each input question.

\paragraph{Retrieval.}
Our retrieval uses LMs for search query generation, relevance rating, and summarization. We independently optimize the prompt choices for search query generation and summarization. The relevance rating prompt is fixed in our system (\autoref{fig:relevance-prompt}).

For search query generation, we evaluate the prompts by retrieving articles with  the generated queries and examining two metrics: (1) the average relevance score across all retrieved articles, and (2) the average relevance score of articles exceeding a relevance threshold of 4 on a 6-point scale. The $2$ high-scoring prompts perform similarly under both metrics and generate queries with little overlap. As a result, we use both prompts (\autoref{fig:both-sq-prompts}) to generate queries and take the union.

For summarization, we run our system end-to-end and pick the top $1$ prompt (\autoref{fig:summarization-prompt}) with respect to the Brier score.

\input{figures/2-calibration-plot}

\paragraph{Reasoning.}
The reasoning system takes a ranked list of article summaries and prompts LMs to make forecasts. We optimize: (1) the ordering criterion of the summaries (by relevance or recency); (2) the number $k$ of article summaries presented to LMs; and (3) the choice of scratchpad instructions to elicit the forecasts.

For efficiency, we optimize them in 2 independent stages:

\begin{itemize}[itemsep=0pt,topsep=0pt]
\item In the first stage, we jointly optimize (1) and (2). Ranking by relevance and setting $k=15$ achieve the lowest average Brier score.

\item In the second stage, we optimize (3) the reasoning prompt. We identify the top 3 prompts out of 15 candidates to elicit 3 predictions from our base model in our system; see \autoref{fig:optimal-reasoning-prompt} for the best one.
\end{itemize}

In optimizing the reasoning system, we test both Claude-2.1 and GPT-4-1106-Preview as candidate models for generating forecasts. GPT-4-1106-Preview consistently yields a 0.01-0.03 lower Brier score. Therefore, our final system elicits predictions from GPT-4-1106-Preview and the fine-tuned GPT-4-0613.

\paragraph{Ensembling.}
We implement 5 ensembling methods, including mean, median, geometric mean, trimmed mean, and a variant of universal self-consistency (USC;  \citet{chen2023universal}). Trimmed mean performs the best in our evaluation; see \autoref{sec:appendix-hp-sweep} for details.

\paragraph{Calibration.} Interestingly, our system is naturally well calibrated (\autoref{fig:calibration_validation}), and we find that standard calibration methods such as binning or isotonic regression do not improve performance.

%% file: figures/fine_tuning_diagram.tex
\begin{figure}[t]
    \centering
    \vspace{-0.9em}
    \includegraphics[width=0.92\textwidth]{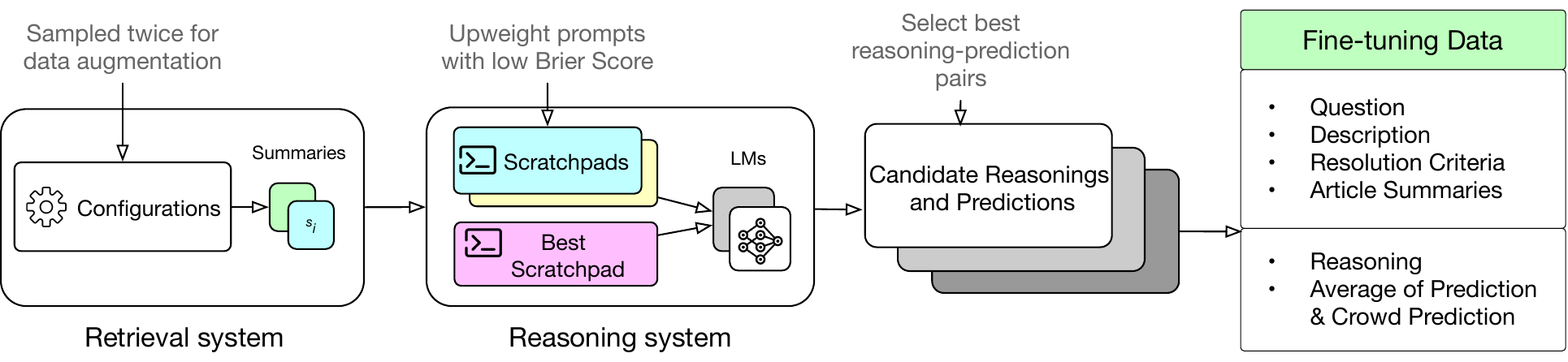}
    \caption{\textbf{Our procedure of generating data for self-supervised training.} For each question, the method generates multiple candidate reasoning-prediction pairs and selects those that outperform human aggregates for fine-tuning.}
    \label{fig:fine-tuning-diagram}
\end{figure}

%% file: figures/2-calibration-plot.tex
\begin{figure}[!t]
    \centering
    \begin{subfigure}[b]{0.32\textwidth}
        \includegraphics[width=\textwidth]{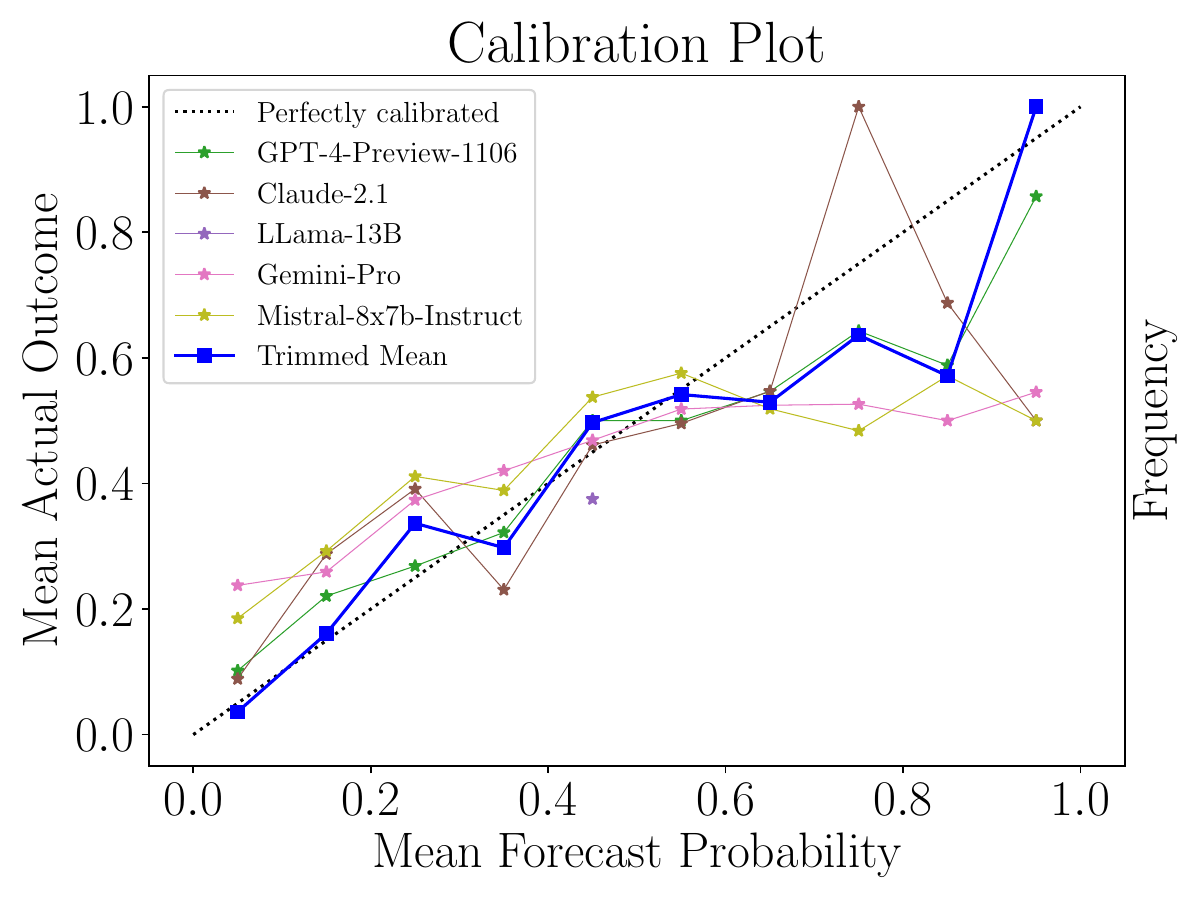}
        \caption{\footnotesize{Calibration of Base Models on Test}}
        \label{fig:calibration_base_eval}
    \end{subfigure}
    \hfill 
    \begin{subfigure}[b]{0.32\textwidth}
        \includegraphics[width=\textwidth]{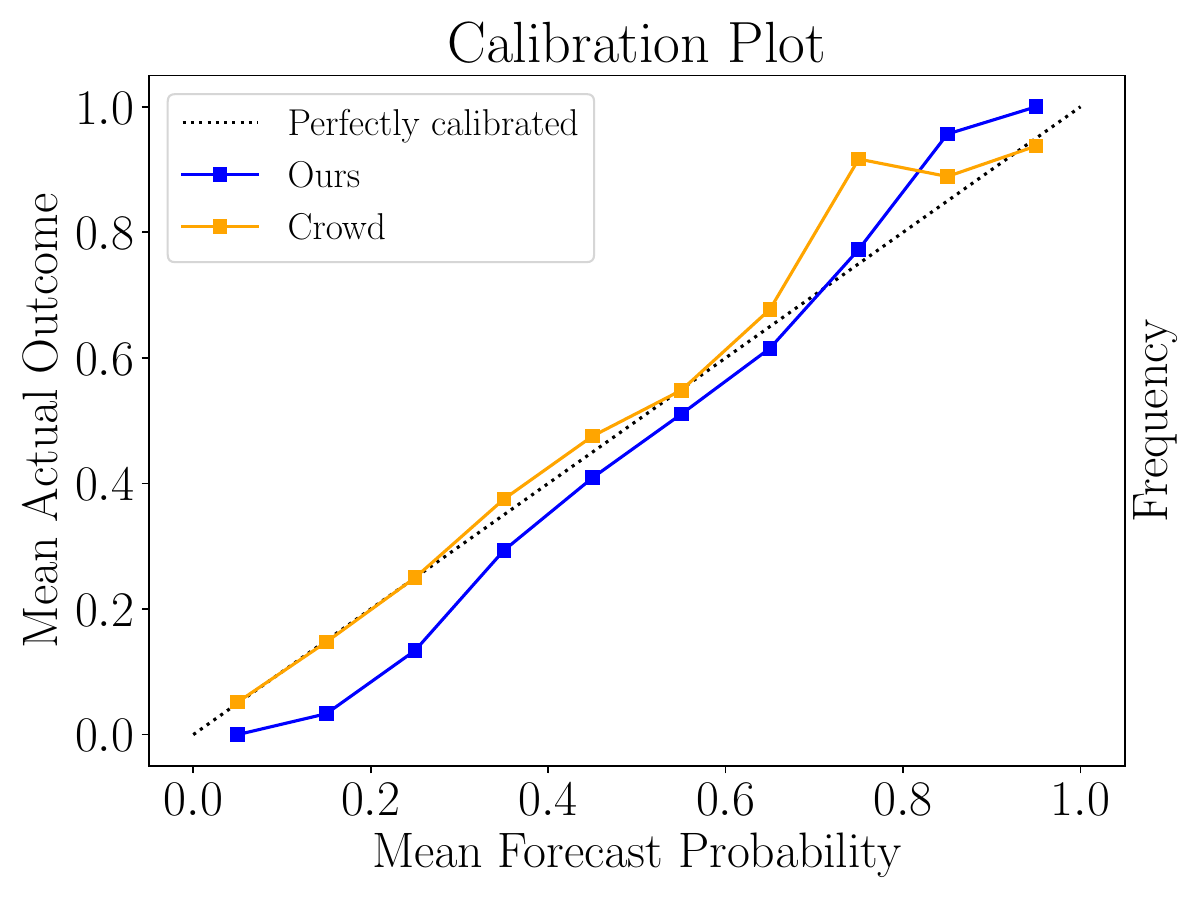}
        \caption{\footnotesize{Calibration of System on Validation}}
        \label{fig:calibration_validation}
    \end{subfigure}
    \hfill 
    \begin{subfigure}[b]{0.32\textwidth}
        \includegraphics[width=\textwidth]{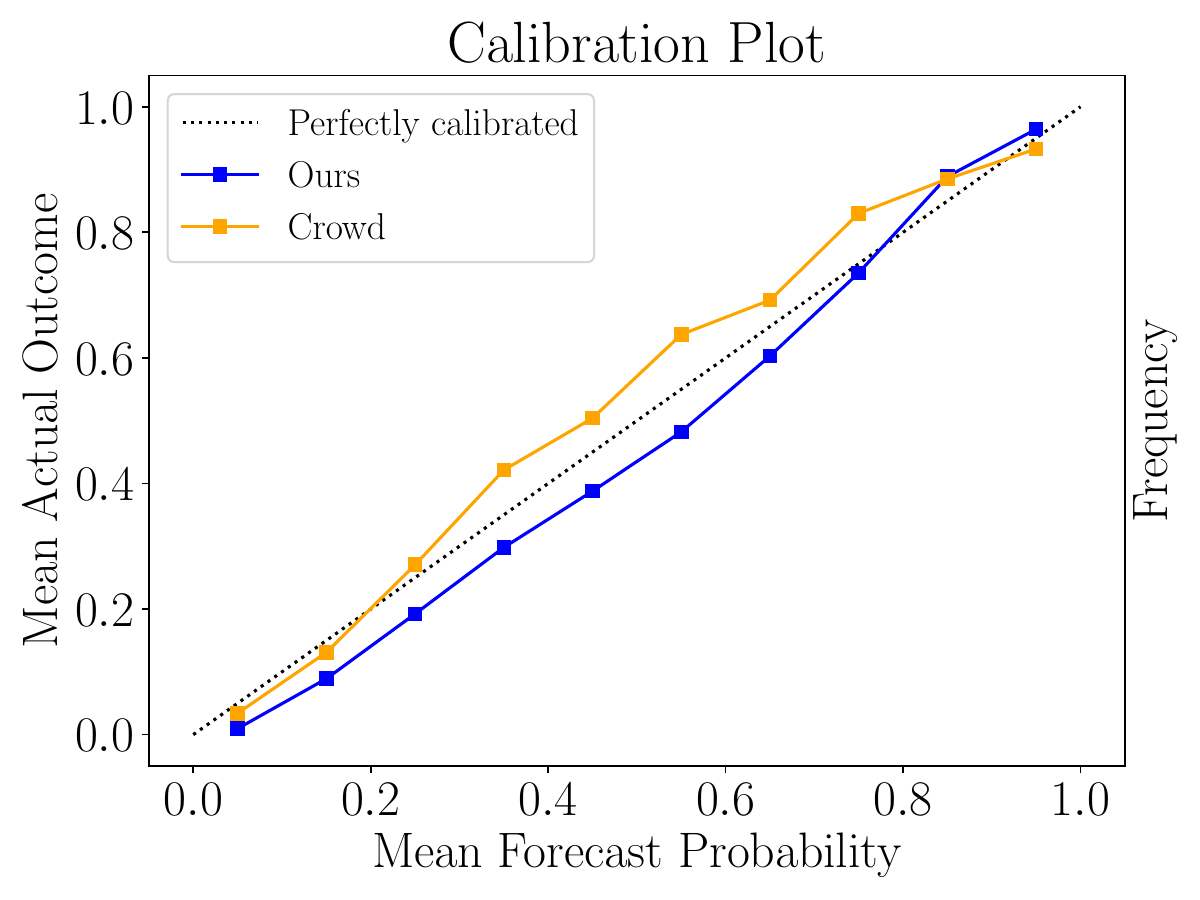}
        \caption{\footnotesize{Calibration of System on Test}}
        \label{fig:calibration_test}
    \end{subfigure}
    \caption{ \textbf{Our system is naturally well calibrated} on both (b) validation and (c) test. The crowd is also well
calibrated, consistent with \citet{zou2022forecasting}'s findings. In contrast, the base models in the zero-shot setting (a) are less calibrated (\autoref{sec:0-shot}). }
    \label{fig:2_plots-calibration}
\end{figure}

%% file: sec/evaluations.tex
\section{Evaluations}
\label{sec:eval}
We evaluate our optimized system  on the test set and find that it comes close to human crowd performance (\autoref{sec:near-sys}). 
Next, we   analyze its strengths and weaknesses (\autoref{subsec:system-strengths}). Motivated by the observations,
we   introduce a relaxed setting, where the system may make forecasts selectively (given its identified strengths), and find that our system surpasses the crowd aggregate (\autoref{sec:relaxed}). Finally, we demonstrate how our system can be used to complement  aggregated human forecasts (\autoref{sec:complement}).

\subsection{System Nears Human Performance}
\label{sec:near-sys}
We first evaluate the Brier score of our end-to-end system on the test set. Note that all hyperparameters were chosen based on the validation set and all test set questions appear temporally after the validation questions, mirroring the setting of a real-time forecasting competition. In addition to the Brier score, we also report accuracy to compare with past work \citep{zou2022forecasting,yan2023autocast}.

\input{tables/system_eval_double_column}

As the main result, our averaged Brier score is $.179$, while the crowd achieves  $.149$, resulting in a difference of $.03$. Our accuracy on the test set is $71.5\%$, whereas the community scores $77.0\%$, resulting in a difference of $5.5\%$. In comparison with the baseline evaluation (\autoref{sec:0-shot}), our system's Brier score ($.179$) significantly outperforms the best baseline model  ($.208$ with GPT-4-1106-Preview)

In prior work, \citet{zou2022forecasting} evaluated their system on 
the forecasting dataset Autocast, which consists of questions from 3 of the platforms we use: Metaculus, INFER, and GJOpen.  They achieved  an accuracy of $65.4\%$ compared to a community baseline of $92.8\%$.  \citet{yan2023autocast}  later improved this to $67.9\%$.
Our results (\autoref{tab:system-evaluation-double}) underscore the significant progress we make in automated forecasting---specifically, we achieve a better   accuracy ($71.5\%$) even though the questions we consider are harder (with a significantly lower crowd accuracy: $77.0\%$).

Further detailed results across different platforms and categories can be found in  \autoref{tab:system-evaluation-double}.
Across categories, our system exhibits noticeable variations: on Sports, our system nearly matches the crowd aggregate, and on Environment \& Energy, it falls much behind. However, we caution against drawing strong conclusions from subcategories, since the sample size is smaller and variation could be due to noise.

Finally, on the test set, we observe again that our system is well calibrated (\autoref{fig:calibration_test}) with RMS calibration error $.42$ (human crowd: $.38$). Interestingly, this is not the case in the baseline evaluations (\autoref{sec:0-shot}), where the models are \textit{not} well calibrated in the zero-shot setting (\autoref{fig:calibration_base_eval}). Through fine-tuning and ensembling, our system improves the calibration of the base models,  without undergoing specific training for calibration.

\subsection{System Strengths and Weaknesses}
\label{subsec:system-strengths}
We next seek to understand our system's strengths and weaknesses. We will investigate these on the validation set, and later use these insights to improve performance on the test set (\autoref{sec:relaxed}). 

We find that our system performs best relative to the crowd on the validation set when (1) the crowd is less confident, (2) at earlier retrieval dates, and (3) when it retrieves many articles. Furthermore, we find that our system is well-calibrated.

First, our system significantly outperforms the crowd when the crowd's predictions express high uncertainty. Specifically, when the crowd's predictions are between $.3$ and $.7$, our Brier score is $.199$ compared to the crowd's $.246$. However, our system underperforms the crowd on questions where they are highly certain, likely because it rarely outputs low probabilities  (\autoref{fig:distribution_of_preds_val}). We hypothesize that this stems from our model's tendency to hedge predictions due to its safety training (see  \autoref{fig:example-kevin} for a qualitative example). Supporting this, our system achieves $7\%$ higher accuracy on questions where the crowd's prediction is within $.05$ of $0$ or $1$, but the Brier score is worse by $.04$.

Next, our system outperforms the crowd on earlier retrieval dates (1, 2, and 3) but not the later ones (4 and 5). Relative to the crowd, our Brier score improves at a slower rate as questions move towards their resolution (\autoref{fig:bs_over_retrieval_val}). This may be due to the aforementioned issue: Our model hedges, even as the evidence becomes more decisive.

With respect to retrieval, our system nears the performance of the crowd when there are at least 5 relevant articles. We further observe that as the number of articles increases, our Brier score improves and surpasses the crowd's (\autoref{fig:num_articles_val}). Intuitively, our system relies on high-quality retrieval, and when conditioned on more articles, it performs better.

Our system is well calibrated on the validation set, with most of the calibration error coming from the system's underconfidence: predictions near $0$ are observed to occur less frequently than anticipated, and similarly, events with predictions close to $1$ also occur at a higher rate than the model suggests (\autoref{fig:calibration_validation}).

\subsection{System Beats Crowd in the Selective Setting}
\label{sec:relaxed}
\input{figures/3_plots_validation}

In real-word forecasting competitions, forecasters do not have to make predictions on every question in the platform at every possible date. Instead, they typically make predictions on questions where they have expertise or interest in and at times that they choose. Therefore, it is natural to leverage our system's strengths and weaknesses and decide accordingly if we should forecast on a retrieval date $k$ for a question $q$.

Leveraging the insights from \autoref{subsec:system-strengths}, we outperform the crowd by making selective forecasts. Specifically, we report the performance when forecasting only under the conditions identified in \autoref{subsec:system-strengths}:
\begin{enumerate} 
    \item \textit{Forecasting only on questions when the crowd prediction falls between $.3$ and $.7$}. Here, our system attains a Brier score of $\mathbf{.238}$ (crowd aggregate: $\mathbf{.240}$). This comprises $51\%$ of forecasts and $56\%$ of questions.
    \item \textit{Forecasting only on earlier retrieval dates (1, 2, and 3)}. Our system's Brier score in this setting is $\mathbf{.185}$ (crowd aggregate: $\mathbf{.161}$). This comprises $66\%$ of forecasts and $100\%$ of questions.
    \item \textit{Forecasting only when the retrieval system provides at least $5$ relevant articles}. Under this condition, our system's Brier score is $\mathbf{.175}$ (crowd aggregate: $\mathbf{.143}$). This makes up $84\%$ of forecasts and $94\%$ of questions.
    \item Under all three conditions, our system attains Brier score $\mathbf{.240}$ (crowd aggregate: $\mathbf{.247}$). This comprises $22\%$ of forecasts and $43\%$ of questions.
\end{enumerate}
The gap in Brier score between our system and the crowd shrinks under each heuristic, except the third one (\autoref{tab:eval-selective-setting}). Under the first heuristic, we outperform the crowd by a small margin ($.238$ vs. $.240$). This is valuable as our system can be used to complement the crowd's prediction when there is greater uncertainty. When all three conditions are jointly met, our system beats the crowd significantly (by more than $1.5$ standard errors in both Brier score and accuracy).

\subsection{System Complements the Crowd}
\label{sec:complement}
Finally, we show that aggregates of our system with the crowd forecasts outperform either one in isolation.

Combining the system's predictions with the crowd using a weighted average---4x weight for the crowd, which we find optimal on the validation set---improves the overall Brier score from $.149$ to $.146$ on the full test set (\autoref{tab:eval-selective-setting}, top row).

Moreover, our system excels under certain criteria (\autoref{subsec:system-strengths}). It is especially useful in these cases to supplement the crowd prediction. We report these results in \autoref{tab:eval-selective-setting} as well, using an unweighted average (instead of the weighted average above). This outperforms the crowd prediction in all cases: For example, the crowd Brier score is $.24$ when the prediction is between $.3$ and $.7$, while the system achieves $.237$. 
%

Finally, beyond direct score improvements, our system can potentially  aid human forecasters by providing  effective news retrieval and novel perspectives in reasoning drawn from LM pre-training knowledge. 
We leave it as a future direction to explore how our system can interactively assist human forecasters.

%% file: tables/system_eval_double_column.tex
\begin{table}[!t]
  \centering
  \resizebox{.5\textwidth}{!}{%
    \begin{tabular}{lccccc}
      \toprule
      \textbf{Category} & \multicolumn{2}{c}{\textbf{Brier Score} $\downarrow$} & \multicolumn{2}{c}{\textbf{Accuracy} $\uparrow$} \\
      \cmidrule(lr){2-3} \cmidrule(lr){4-5}
      & \textbf{Ours} & \textbf{Crowd} & \textbf{Ours} & \textbf{Crowd} \\
      \midrule
      Science \& Tech & $.143_{.011}$ & $.114_{.011}$ & $82.2_{2.7}$ & $84.3_{2.6}$ \\
Healthcare \& Biology & $.074_{.015}$ & $.063_{.020}$ & $93.8_{4.3}$ & $90.6_{5.2}$ \\
Economics \& Business & $.198_{.007}$ & $.147_{.009}$ & $68.8_{2.1}$ & $78.3_{1.9}$ \\
Politics \& Governance & $.172_{.006}$ & $.145_{.007}$ & $72.6_{1.4}$ & $78.2_{1.3}$ \\
Education \& Research & $.163_{.024}$ & $.129_{.024}$ & $80.6_{6.7}$ & $77.8_{7.0}$ \\
Arts \& Recreation & $.221_{.010}$ & $.146_{.010}$ & $62.4_{2.5}$ & $76.9_{2.2}$ \\
Security \& Defenses & $.174_{.008}$ & $.129_{.009}$ & $71.0_{2.1}$ & $78.4_{1.9}$ \\
Sports & $.175_{.004}$ & $.171_{.005}$ & $73.0_{1.3}$ & $73.1_{1.3}$ \\
      \midrule
    \textbf{All Categories} & $\mathbf{.179_{.003}}$ & $\mathbf{.149_{.003}}$ & $\mathbf{71.5_{.7}}$ & $\mathbf{77.0_{.7}}$ \\
      \bottomrule
    \end{tabular}%
  }%
  \quad
    \resizebox{.45\textwidth}{!}{%
    \begin{tabular}{lccccc}
      \toprule
      \textbf{Platform} & \multicolumn{2}{c}{\textbf{Brier Score} $\downarrow$} & \multicolumn{2}{c}{\textbf{Accuracy} $\uparrow$} \\
      \cmidrule(lr){2-3} \cmidrule(lr){4-5}
      & \textbf{Ours} & \textbf{Crowd} & \textbf{Ours} & \textbf{Crowd} \\
      \midrule
      Metaculus & $.134_{.005}$ & $.104_{.005}$ & $80.3_{1.2}$ & $86.6_{1.1}$ \\
GJOpen & $.193_{.011}$ & $.157_{.013}$ & $67.9_{3.4}$ & $72.6_{3.2}$ \\
INFER & $.247_{.053}$ & $.310_{.086}$ & $60.0_{13.1}$ & $53.3_{13.3}$ \\
Polymarket & $.172_{.005}$ & $.127_{.006}$ & $73.6_{1.3}$ & $79.9_{1.1}$ \\
Manifold & $.219_{.004}$ & $.200_{.005}$ & $63.6_{1.3}$ & $67.9_{1.3}$ \\
      \midrule
      \textbf{All Platforms} & $\mathbf{.179_{.003}}$ & $\mathbf{.149_{.003}}$ & $\mathbf{71.5_{.7}}$ & $\mathbf{77.0_{.7}}$ \\
      \bottomrule
    \end{tabular}%
  }
  \caption{\textbf{Results of system evaluation} by category \textbf{(left)} and by platform \textbf{(right)}. Subscript numbers are $1$ standard error. Averaged across all retrieval dates, our optimal system, as described in \autoref{sec:system}, achieves $.179$ Brier score (human crowd: $.149$) and accuracy $.715$ (human crowd: $.770$).}
  \label{tab:system-evaluation-double}
\end{table}

%% file: figures/3_plots_validation.tex
\begin{figure}[!t]
    \centering
    \begin{subfigure}[b]{0.32\textwidth}
        \includegraphics[width=\textwidth]{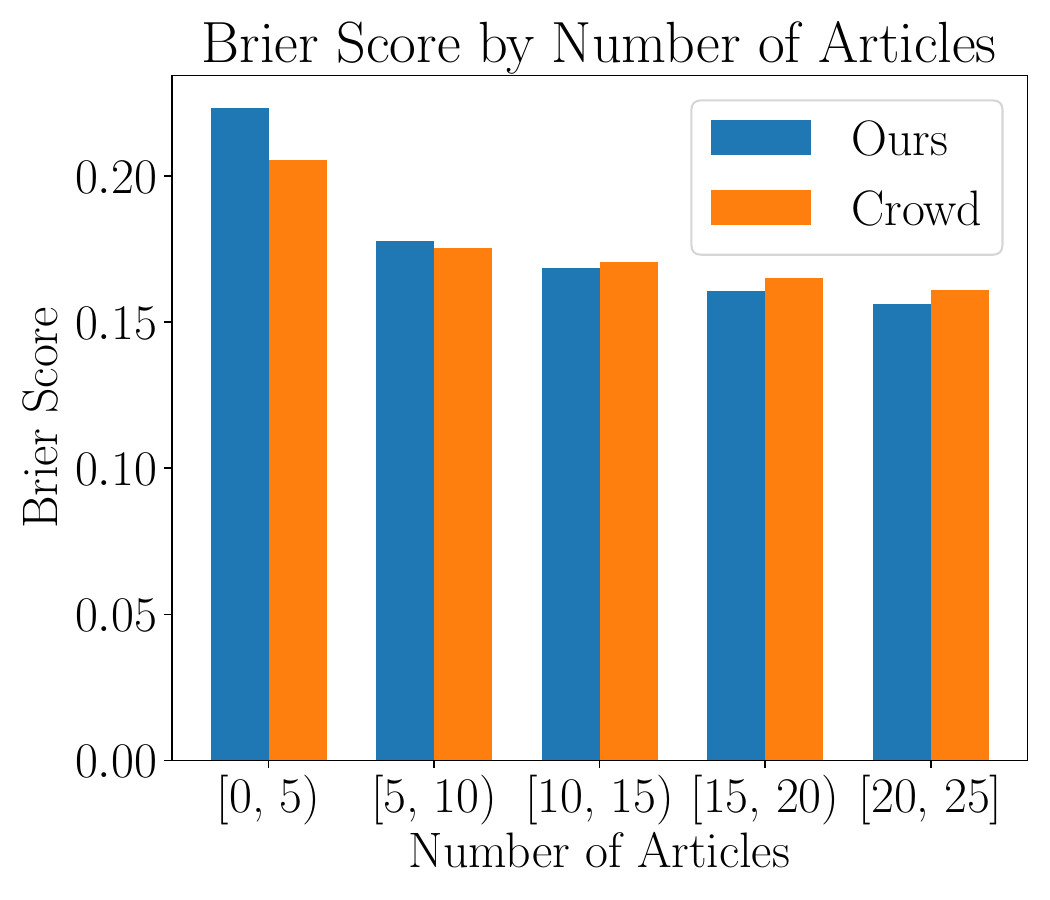}
        \caption{Brier Score w.r.t. $\#$ of Articles}
        \label{fig:num_articles_val}
    \end{subfigure}
    \hfill 
    \begin{subfigure}[b]{0.32\textwidth}
        \includegraphics[width=\textwidth]{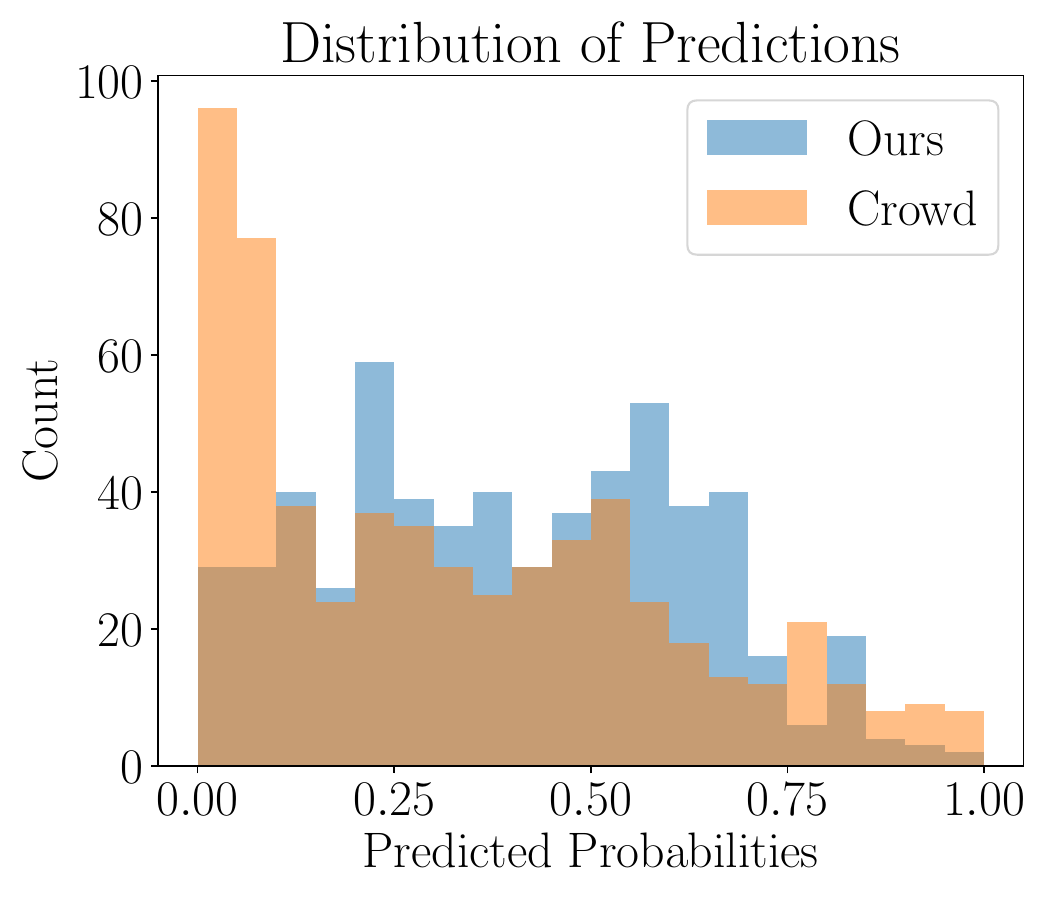}
        \caption{Distribution of predictions.}
        \label{fig:distribution_of_preds_val}
    \end{subfigure}
    \hfill 
    \begin{subfigure}[b]{0.32\textwidth}
        \includegraphics[width=\textwidth]{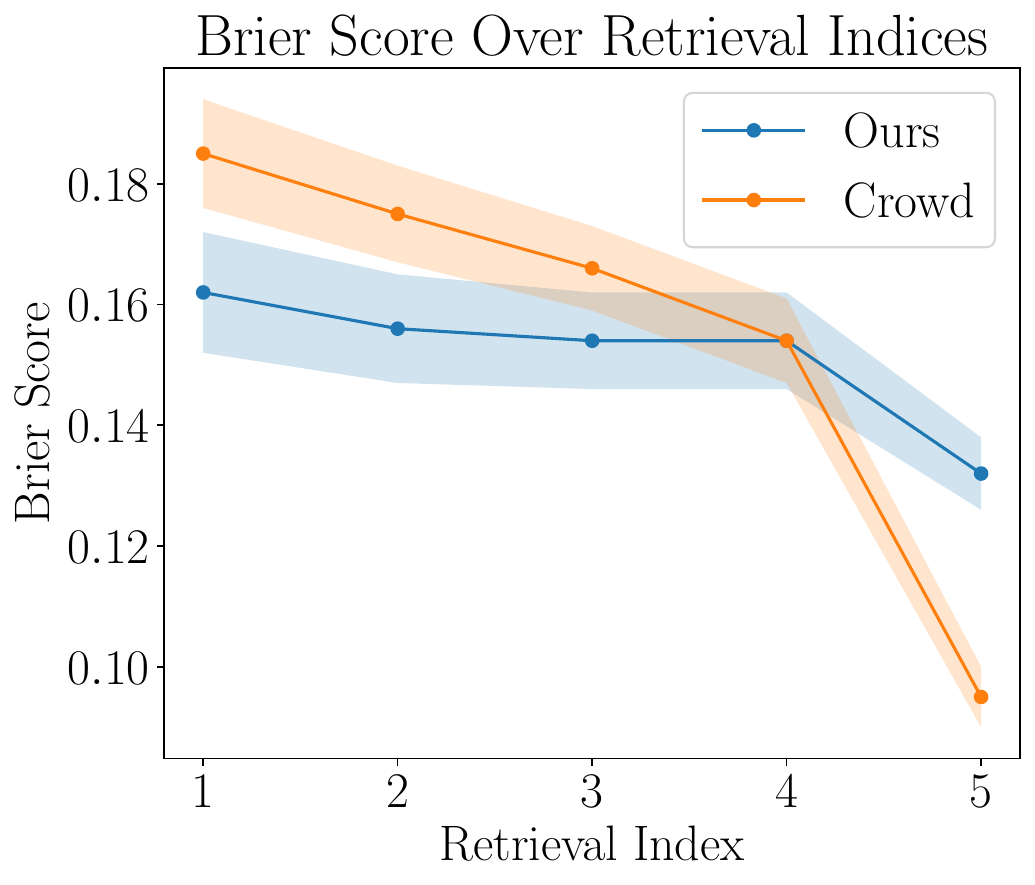}
        \caption{Brier score over retrieval indices.}
        \label{fig:bs_over_retrieval_val}
    \end{subfigure}
    \caption{\textbf{System strengths}. Evaluating on the validation set, we note: \textbf{(a)} When provided enough relevant articles, our system outperforms the crowd. \textbf{(b)} For questions where the crowd is unsure (predictions between $.3$ and $.7$), we outperform them (Brier score $.199$ vs. $.246$). However, the crowd outperforms our system on questions where they are highly confident, e.g. predicting less than .05. \textbf{(c)} Our system's Brier score is better at the earlier retrieval dates. Finally, our system is well-calibrated (c.f. \autoref{fig:calibration_validation}). }
    \label{fig:4_plots}
\end{figure}

%% file: sec/ablations.tex
\section{Ablations}
\label{sec:ablate-main}
We conduct $3$ ablation studies. The first validates that our performance is not solely due to the power of GPT-4. The last two show the benefits of our retrieval and fine-tuning methods.
\input{tables/one_table_ablation}

\paragraph{Fine-tuning a less capable model.}
To demonstrate that our system's performance does not hinge on the ability of the base model (i.e., GPT-4), we fine-tune GPT-3.5 on all our fine-tuning data (13,253 samples).

We replace fine-tuned GPT-4 in our system with fine-tuned GPT-3.5, and evaluate using the same methodology as in \autoref{sec:near-sys}. We find here that our Brier score is only slightly worse: $.182$ compared to the previous score of $.179$.

\paragraph{No fine-tuning.} 
To demonstrate the gain from fine-tuning (\autoref{subsec:fine_tuning}), we evaluate our optimal system, except we only use base GPT-4-Preview-1106 as the reasoning model. 

In this setup, the ablated system achieves a Brier score of $.186$, which increased on the original score by $.007$. 

Overall, the results suggest that fine-tuning the reasoning model yields a significant boost to our system's performance.
 
\paragraph{No fine-tuning and no retrieval.}
We evaluate our optimal system without any news retrieval and using the base GPT-4-1106-Preview model. The ablated system attains a Brier score of $.206$.

Recall that in our baseline evaluation (\autoref{sec:0-shot}), the lowest Brier score attained by any model is $.208$. Our ablated system essentially deteriorates to this baseline level. Indeed, without any fine-tuning or retrieval, the only expected advantage of our system over the baseline evaluation setup is its reasoning prompt, found through searching  a set of candidate prompts (\autoref{sec:optimization}). The experiment suggests that this gives fairly a minor improvement.

%% file: tables/one_table_ablation.tex
\begin{table}[!t]
    \centering
    \begin{tabular}{lcccccccc}
      \toprule
      \textbf{Criteria} & \multicolumn{2}{c}{\textbf{Brier Score} $\downarrow$} & \multicolumn{2}{c}{\textbf{\% Accuracy} $\uparrow$} \\
      \cmidrule(lr){2-3} \cmidrule(lr){4-5} 
      & \textbf{Ours} & \textbf{Aggregate} & \textbf{Ours} & \textbf{Aggregate} \\
      \midrule
      \textbf{Full System} & $.179_{.003}$ & $\mathbf{.146_{.002}}$ & $71.5_{.7}$ &  $\mathbf{77.8_{.6}}$ \\
      \textbf{Fine-tuned GPT-4-0613} & $.182_{.002}$ & $\mathbf{.146_{.002}}$ & $70.7_{.7}$ & $\mathbf{77.4_{.6}}$ \\
      \textbf{Fine-tuned GPT-3.5 \& Base GPT-4} & $.181_{.002}$ & $\mathbf{.147_{.002}}$ & $70.9_{.7}$ & $\mathbf{77.4_{.6}}$ \\
      \textbf{Fine-tuned GPT-3.5} & $.183_{.002}$ & $\mathbf{.146_{.002}}$ & $71.5_{.7}$ & $\mathbf{77.4_{.6}}$ \\
      \textbf{Base GPT-4} & $.186_{.002}$ & $\mathbf{.148_{.002}}$ & $70.6_{.7}$ & $\mathbf{77.1_{.6}}$ \\
      \textbf{Base GPT-4; no IR} & $.206_{.002}$ & $.150_{.002}$ & $66.6_{.7}$ & $76.9_{.6}$ \\
      \bottomrule
    \end{tabular}
    \caption{\textbf{Ablation study results.} The crowd Brier score and accuracy are .146 and 77.0\%, respectively. ``Aggregate'' indicates the weighted average of our system with the crowd prediction. Our full system uses fine-tuned GPT-4-0613 and base GPT-4-1106-Preview \textbf{(row 1)}. The system yields similar performance with fine-tuned GPT-3.5 \textbf{(rows 3--4)}. Our system exhibits poorer performance without a fine-tuned reasoning model \textbf{(row 5)}, and further declines with neither retrieval nor a fine-tuned reasoning model \textbf{(row 6)}. Subscript numbers represent one standard error. We bold entries that surpass the crowd aggregate.}
  \label{tab:one-ablation-table}
\end{table}

%% file: sec/conclusion.tex
\section{Conclusion}
\label{sec:discussion}
Our work presents the first ML system that can forecast at near human levels. We develop a novel retrieval mechanism that uses a LM to determine which information to source and how to evaluate its relevance. We also give a self-supervised fine-tuning method to generate reasonings with accurate predictions.

To facilitate further research, we release our dataset: the largest and most recent forecasting dataset compiled from 5 real-world forecasting competitions. We discuss a few opportunities to improve these systems further.

\paragraph{Iterative self-supervision.} With a larger training corpus, our self-supervised fine-tuning approach can be used for iterative self-improvement. Specifically, after fine-tuning a model on its previous optimal predictions and reasonings, we can generate more fine-tuning data by using the same model again, which can be repeated until training data is exhausted.

\paragraph{Data.} While our forecasting benchmark is a good initial corpus to train a system, we believe that it is possible to use  LMs with later training cut-offs to teach an earlier LM. This could be done by using later LMs to generate questions  it knows the answer to but an earlier LM does not (postdiction). In addition, while we source questions from forecasting platforms, it is possible to collect historical data in the wild and re-formulate them as forecasting questions, leading to a larger training set.

\paragraph{Domain-adaptive training.} In \autoref{sec:knowledge-eval}, we observe that in the baseline evaluations, the Brier scores across categories are correlated with models' pre-training knowledge. This suggests that we may be able to specialize models to areas of particular interests by fine-tuning them on domain knowledge.

\paragraph{LMs get better at forecasting naturally.} We observe that as LMs improve, they naturally also become better at forecasting. In particular, in \autoref{sec:0-shot}, we see that newer generations of models forecast  better than older ones. For example, GPT-4-1106, released in 2023, outperforms GPT-4-0613, released in 2021, by $.02$ with respect to the Brier score. If we were to have fine-tuned the more recent model, we would expect better performance.

At a high level, our results suggest that in the near future, LM-based  systems may be able to generate accurate forecasts at the level of competitive human forecasters.  We hope that our work paves the way for automated, scalable forecasting that can help to inform institutional decision making.


%% file: appendix/models.tex
\section{Details about Models and Knowledge Accuracy}
\subsection{Models}
\label{sec:appendix-models}
We give a list of detailed information of the models we use below. The weights of the open models are available publicly on Hugging Face, and we primarily use Together AI's serving API to access them. All cut-offs are based on official statements.

\input{tables/model_source}

%% file: tables/model_source.tex
\begin{table}[h!]
  \centering
  \small
  \begin{tabular}{@{}l l l l r@{}}
    \toprule
    \textbf{Model} & \textbf{Source} & \textbf{Open Weights} & \textbf{Knowledge Cut-off} & \multicolumn{1}{l}{\textbf{Evaluation Cost}} \\
    \midrule
    GPT-4-1106-Preview & OpenAI & No & Apr 2023 & \$0.01/1K tokens \\
    GPT-4 (GPT-4-0613) & OpenAI & No & Sep 2021 & \$0.03/1K tokens \\
    GPT-3.5-Turbo-Instruct & OpenAI & No & Sep 2021 & \$0.0015/1K tokens \\
    GPT-3.5-Turbo-1106 & OpenAI & No & Sep 2021 & \$0.001/1K tokens \\
    Claude-1 & Anthropic & No & Dec 2022 & \$0.024/1K tokens \\
    Claude-2 & Anthropic & No & Dec 2022 & \$0.024/1K tokens \\
    Claude-2.1 & Anthropic & No & Dec 2022 & \$0.024/1K tokens \\
    Llama-2-7B-Chat & Meta & Yes & Sep 2022 & \$0.0002/1K tokens \\
    Llama-2-13B-Chat & Meta & Yes & Sep 2022 & \$0.00025/1K tokens \\
    Llama-2-70B-Chat & Meta & Yes & Sep 2022 & \$0.0009/1K tokens \\
    Mistral-7B-Instruct & Mistral AI & Yes & \textit{Unknown} & \$0.0002/1K tokens \\
    Mistral-8x7B-Instruct & Mistral AI & Yes &\textit{Unknown} & \$0.0002/1K tokens \\
    Mixtral-8x7B-DPO & NousResearch & Yes &\textit{Unknown}  & \$0.0002/1K tokens \\
    YI-34B-Chat & 01.AI & Yes & June 2023 & \$0.000776/1K tokens \\
    Gemini-Pro & Google & No & Early 2023 & \$0.0005/1K characters \\
    \bottomrule
  \end{tabular}
  \caption{\textbf{Overview of the LMs we evaluate}: A breakdown of the LMs used in our study, including their sources, availability of weights, knowledge cut-off dates, and evaluation costs. The evaluation costs of the open-weight models are based on Together AI's pricing. The knowledge cut-off of Gemini-Pro is claimed to be early 2023 ($\sim$ April 2023).  We are not aware of the exact knowledge cut-offs of the Mistral series, as it is not publicly reported.}
  \label{tab:models} 
\end{table} 

\subsection{Testing Potential Leakage from Post-training}
GPT-4-1106-Preview and GPT-3.5-Turbo-1106, the two models we use in our system, were released in November, 2023. We find no evidence that     the post-training phase leaks further information after their knowledge cut-offs (April, 2023 and January, 2021).  As a test, we manually query the model on $20$ major events in June, 2023--September, 2023\footnote{sourced from \url{https://www.onthisday.com/events/date/2023/}.}, such as ``Who won the 2023 Turkish presidential election?''. For all $20$ questions, both models either claim  no knowledge or simply hallucinate.

As a sanity check, we also prompt GPT-4-1106-Preview to answer another $20$ questions about events during November, 2022--January, 2023, prior to its  knowledge cut-off, such as ``Which team won the 2022 FIFA World Cup Final?''. The model answers all of them correctly.

\subsection{Crowd Predictions}
\label{sec:crowd-pred}
On any given question, each platform computes a community prediction that aggregates all individual forecasts. The prediction is dynamically updated and recorded as the forecasts are made. We source the records directly from the platforms (instead of computing them from scratch using the individual forecasts). For binary questions, we provide more details on the aggregation mechanisms as follows.
\begin{itemize}
    \item  On Metaculus, for a given question,  each prediction of a forecaster is marked by $t$  (starting at $1$), from their earliest   prediction to the latest.  
    The platform computes the crowd prediction of the question by weighted median. The weight of the $t$th forecast from an individual forecaster is   $e^{\sqrt{t}}$, so the more recent forecasts receive higher weights. 
    We remark that  the platform also publishes another aggregated forecast called ``Metaculus prediction'' (which we do not use or compare with in this paper). This differs from the crowd prediction described above and   is computed via a proprietary algorithm.
    \item GJOpen computes the crowd predictions by the mean of the most recent 40\% of the  forecasts from each forecaster.
    \item  INFER initializes the crowd   prediction to be the mean of all individual forecasts. As the question progresses, it reweights the forecasts, for example, by ``putting more weight on the forecasts of individuals with the best track record.''\footnote{\url{https://www.infer-pub.com/frequently-asked-questions}}   Exact details on the aggregation mechanisms are not found on their website.
    \item Manifold and Polymarket are prediction markets, where the community predictions are the prices (between 0 and 1).  The prices are adjusted  by their automated market makers, as bets are made.
    The   mechanisms are variants of  constant-product market makers \citep{hanson2007logarithmic, zhang2018formal}; see \cite{Polymarket, Notion} for more details. 
\end{itemize}

%% file: appendix/base_eval.tex
\section{Details about Base Evaluations}
\label{sec:base-eval}
In this section, we provide experimental details on our baseline evaluations (\autoref{sec:0-shot}). 

\subsection{Evaluation Method}
For both zero-shot and scratchpad prompting, we conduct basic prompt optimization by by crafting 5 candidate zero-shot prompts and 4 candidate scratchpad prompts. We evaluate each prompt on the validation set by comparing Brier scores. Specifically, we randomly select $200$ questions from the validation set and calculate the mean Brier scores across the $14$ LMs under consideration.
\begin{itemize}
    \item 
The best zero-shot prompt achieves an average Brier score of 0.246, outperforming the others, which score 0.261, 0.276, 0.279, and 0.252, respectively. 
\item 
For scratchpad, all prompts yield  similar Brier scores. 
We observe that potentially due to safety training, models can sometimes refuse to answer forecasting questions by simply claiming ``I don't know''.
Therefore, we use the number of ``refuse to answer'' responses as the deciding metric. The winning scratchpad prompt averages 88 ``refuse to answer'' responses, while the others average 106, 93, and 94, respectively.
\end{itemize}
The best zero-shot and scratchpad prompts are shown in \autoref{fig:zero-shot-base-eval-prompt} and \autoref{fig:scratchpad-base-eval-prompt}. In both prompting styles,  models are only provided with the question, background, resolution criterion, and question's open and close dates (\texttt{\small{date\_begin}} and \texttt{\small date\_end}). All the data are sourced  from the forecasting platforms and publicly available on the question page to human forecasters. We do no additional news retrieval.

\input{figures/base_eval_zero_shot_prompt}
\input{figures/base_eval_scratch_pad_prompt}

Finally, we use  the best prompt of each prompting strategy to forecast on each question in the test set. In  \autoref{sec:0-shot}, we find that none of the models are naturally good at forecasting. We provide the full results next in \autoref{sec:full-base-eval}.
\clearpage

\subsection{Baseline Evaluation Results}
\label{sec:full-base-eval}
We now give the full results of our baseline evaluation (\autoref{sec:0-shot}) in \autoref{tab:model_comparison}.
\input{tables/base_eval_zero_shot_and_scratch_pad}

\subsection{Knowledge Evaluation by Category}
\label{sec:knowledge-eval}

We present an evaluation of model's knowledge  about resolved questions on past events and notice variations in performance across categories. To investigate further, we analyzed each model's zero-shot Brier score on the test set by category. This analysis showed a correlation between models' knowledge on the training and validation sets and their Brier scores on the test set across categories. This suggests that domain-adaptive training could be used to improve model performance in categories where its existing knowledge is limited.

First, we assessed pre-trained language model knowledge across categories by evaluating their ability to answer resolved forecasting questions from the train and validation sets. See \autoref{tab:knowledge-accuracy-cate} for the results and \autoref{fig:model-knowledge-eval-prompt} for the knowledge prompt.

\input{tables/base_eval_knowledge_by_category}
\input{figures/base_eval_knowledge_prompt}

We noticed variations in knowledge accuracy across categories. To dig deeper, we analyze the zero-shot Brier score on the test set in \autoref{tab:zero-shot-cate} and assess if there is a correlation between knowledge accuracy on the training and validation sets and zero-shot Brier score on the test set in \autoref{tab:knowledge-zero-shot-correlation}.

\input{tables/base_eval_zero_shot_by_category}

\paragraph{The potential for domain-adaptive training.} 
We calculate the correlation between the models' knowledge accuracy and their Brier scores of the zero-shot evaluation. Notably, in the Politics \& Governance, Arts  \& Recreation, and Education \& Research categories, there exists a strong negative correlation. See the below \autoref{tab:knowledge-zero-shot-correlation} for the correlation table. This negative correlation is expected because a higher knowledge accuracy should intuitively correspond to a lower Brier score. As a direction for future research, we propose that domain-adaptive training could be employed to enhance forecasting performance in specific categories.

\input{tables/correlation_between_knowledge_and_zero_shot}

%% file: figures/base_eval_zero_shot_prompt.tex
\begin{figure}[ht!]
    \centering
\begin{mdframed}
You are an expert superforecaster, familiar with the work of Tetlock and others. Make a prediction of the probability that the question will be resolved as true. You MUST give a probability estimate between 0 and 1 UNDER ALL CIRCUMSTANCES. If for some reason you can't answer, pick the base rate, but return a number between 0 and 1.
\\

Question:
\{question\}

Question Background: \{background\}

Resolution Criteria: \{resolution\_criteria\}
\\

Today's date: \{date\_begin\}

Question close date: \{date\_end\}
\\ 

Output your answer (a number between 0 and 1) with an asterisk at the beginning and end of the decimal. Do not output anything else.

Answer: \{\{ Insert answer here \}\}
\end{mdframed}
 \caption{\textbf{The simple zero-shot prompt used for  baseline evaluations}. No retrieval is performed. The prompt simply asks the model to make a prediction on a given question from the test set.  We add the directive ``You MUST ... UNDER ALL CIRCUMSTANCES'' to push the model to answer the question, which in some cases it refuses to, potentially due to safety training. See \autoref{sec:0-shot} for results and \autoref{sec:base-eval} for more details.}
    \label{fig:zero-shot-base-eval-prompt}
\end{figure}

%% file: figures/base_eval_scratch_pad_prompt.tex
\begin{figure}[ht!]
    \centering
\begin{mdframed}
Question:
\{question\}

Question Background:\{background\}

Resolution Criteria:\{resolution\_criteria\}
\\

Today's date: \{date\_begin\}

Question close date: \{date\_end\}
\\ 

Instructions:\\
1. Provide reasons why the answer might be no.

\{\{ Insert your thoughts \}\}
\\

2. Provide reasons why the answer might be yes.

\{\{ Insert your thoughts \}\}
\\

3. Aggregate your considerations.

\{\{ Insert your aggregated considerations \}\}
\\

4. Output your answer (a number between 0 and 1) with an asterisk at the beginning and end of the decimal.

\{\{ Insert your answer \}\}
\end{mdframed}
 \caption{\textbf{The scratchpad prompt used for baseline evaluations.} No retrieval is performed. The prompt asks the model to make a prediction on a given question from the test set, after making considerations for yes and no. See \autoref{sec:0-shot} for results and \autoref{sec:base-eval} for more details.}
    \label{fig:scratchpad-base-eval-prompt}
\end{figure}

%% file: tables/base_eval_zero_shot_and_scratch_pad.tex
\begin{table}[ht!]
  \centering
  \begin{tabular}{@{}l r r@{}}
    \toprule
    \textbf{Model} & \textbf{Zero-shot} & \textbf{Scratchpad} \\
    \midrule
    GPT-3.5-Turbo & $0.237$ $(0.014)$ & $0.257$ $(0.009)$ \\
    GPT-3.5-Turbo-1106 & $0.274$ $(0.016)$ & ${0.261}$ $(0.010)$ \\
    GPT-4 (GPT-4-0613) & $0.219$ $(0.013)$ & $0.222$ $(0.009)$ \\
    GPT-4-1106-Preview & $\mathbf{0.208}$ $(0.013)$ & $\mathbf{0.209}$ $(0.012)$ \\
    Llama-2-7B & $0.353$ $(0.020)$ & ${0.264}$ $(0.011)$ \\
    Llama-2-13B & $0.226$ $(0.009)$ & $0.268$ $(0.008)$ \\
    Llama-2-70B & $0.283$ $(0.014)$ & ${0.282}$ $(0.011)$ \\
    Mistral-7B-Instruct & $0.237$ $(0.018)$ & $0.243$ $(0.008)$ \\
    Mistral-8x7B-Instruct & $0.238$ $(0.018)$ & $0.238$ $(0.010)$ \\
    Mixtral-8x7B-DPO & $0.260$ $(0.022)$ & ${0.248}$ $(0.010)$ \\
    Yi-34B-Chat & $0.238$ $(0.012)$ & $0.241$ $(0.009)$ \\
    Claude-2 & $0.220$ $(0.013)$ & ${0.219}$ $(0.014)$ \\
    Claude-2.1 & $0.220$ $(0.013)$ & ${0.215}$ $(0.014)$ \\
    Gemini-Pro & $0.243$ $(0.019)$ & ${0.230}$ $(0.007)$ \\
    \bottomrule
  \end{tabular}
  \caption{\textbf{Zero-shot and scratchpad Brier scores} on the test set:     Brier scores under zero-shot or scratchpad prompts, with 2 standard error (SE) values. Lower is better.  Random baseline: 0.250; human crowd: 0.149. All models fall significantly far from human aggregate.}
  \label{tab:model_comparison} 
\end{table}

%% file: tables/base_eval_knowledge_by_category.tex
\begin{table}[ht!]
  \centering
  \resizebox{\textwidth}{!}{%
  \begin{tabular}{@{}lccccccccccc@{}}
    \toprule
    \Large Model & \Large Arts \& Recreation & \Large Economics \& Business & \Large Education \& Research & \Large Environment \& Energy & \Large Healthcare \& Biology & \Large {Politics \& Governance} & \Large Science \& Tech & \Large Security \& Defense & \Large Social Sciences & \Large Sports & \Large Other \\
    \midrule 
    \Large GPT-3.5-Turbo & \Large 0.411 & \Large 0.323 & \Large 0.25 & \Large 0.314 & \Large 0.419 & \Large {0.328} & \Large 0.387 & \Large 0.314 & \Large 0.462 & \Large 0.365 & \Large 0.107 \\
    \Large GPT-3.5-Turbo-1106 & \Large 0.196 & \Large 0.195 & \Large 0.375 & \Large 0.229 & \Large 0.262 & \Large {0.278} & \Large 0.247 & \Large 0.286 & \Large 0.154 & \Large 0.166 & \Large 0.25 \\
    \Large GPT-4 (GPT-4-0613) & \Large 0.083 & \Large 0.114 & \Large 0.125 & \Large 0.22 & \Large 0.157 & \Large {0.349} & \Large 0.196 & \Large 0.279 & \Large 0.077 & \Large 0.04 & \Large 0.071 \\
    \Large GPT-4-1106-Preview & \Large 0.094 & \Large 0.142 & \Large 0.125 & \Large 0.153 & \Large 0.144 & \Large {0.391} & \Large 0.227 & \Large 0.207 & \Large 0.0 & \Large 0.234 & \Large 0.0 \\
    \Large Llama-2-7B & \Large 0.042 & \Large 0.069 & \Large 0.156 & \Large 0.203 & \Large 0.284 & \Large {0.046} & \Large 0.067 & \Large 0.033 & \Large 0.0 & \Large 0.05 & \Large 0.071 \\
    \Large Llama-2-13B & \Large 0.102 & \Large 0.181 & \Large 0.156 & \Large 0.288 & \Large 0.288 & \Large {0.21} & \Large 0.247 & \Large 0.163 & \Large 0.231 & \Large 0.189 & \Large 0.036 \\
    \Large Llama-2-70B & \Large 0.143 & \Large 0.175 & \Large 0.344 & \Large 0.322 & \Large 0.266 & \Large {0.243} & \Large 0.384 & \Large 0.115 & \Large 0.077 & \Large 0.075 & \Large 0.107 \\
    \Large Mistral-7B-Instruct & \Large 0.011 & \Large 0.024 & \Large 0.0 & \Large 0.034 & \Large 0.022 & \Large {0.05} & \Large 0.054 & \Large 0.007 & \Large 0.0 & \Large 0.018 & \Large 0.0 \\
    \Large Mistral-8x7B-Instruct & \Large 0.019 & \Large 0.045 & \Large 0.094 & \Large 0.051 & \Large 0.066 & \Large {0.071} & \Large 0.049 & \Large 0.04 & \Large 0.0 & \Large 0.027 & \Large 0.0 \\
    \Large Mixtral-8x7B-DPO & \Large 0.004 & \Large 0.084 & \Large 0.031 & \Large 0.051 & \Large 0.087 & \Large {0.041} & \Large 0.054 & \Large 0.014 & \Large 0.0 & \Large 0.01 & \Large 0.0 \\
    \Large YI-34B-Chat & \Large 0.423 & \Large 0.552 & \Large 0.625 & \Large 0.593 & \Large 0.555 & \Large {0.63} & \Large 0.588 & \Large 0.738 & \Large 0.538 & \Large 0.624 & \Large 0.536 \\
    \Large Claude-2 & \Large 0.14 & \Large 0.205 & \Large 0.219 & \Large 0.254 & \Large 0.245 & \Large {0.446} & \Large 0.296 & \Large 0.134 & \Large 0.154 & \Large 0.392 & \Large 0.071 \\
    \Large Claude-2.1 & \Large 0.136 & \Large 0.205 & \Large 0.219 & \Large 0.246 & \Large 0.249 & \Large {0.446} & \Large 0.294 & \Large 0.136 & \Large 0.077 & \Large 0.395 & \Large 0.071 \\
    \Large Gemini-Pro & \Large 0.155 & \Large 0.425 & \Large 0.188 & \Large 0.415 & \Large 0.314 & \Large {0.425} & \Large 0.356 & \Large 0.545 & \Large 0.077 & \Large 0.35 & \Large 0.25 \\
    \bottomrule
  \end{tabular}%
  }
  \caption{\textbf{Comparison of knowledge accuracy across categories and models} on the train and validation sets. We list the knowledge accuracy of all base models with respect to all categories in the train  and validation set.}
  \label{tab:knowledge-accuracy-cate}
\end{table}

%% file: figures/base_eval_knowledge_prompt.tex
\begin{figure}[ht!]
    \centering
    \begin{mdframed}
        Question: \{question\}

        The question was posed on \{date\_begin\} and closed on \{date\_end\}.

        Instructions: 
        \begin{itemize}
            \item Please output "1" if the answer is "Yes", "0" if the answer is "No" or "IDK" if you don't know the answer. Do not return anything else.
            \item Do not guess.
        \end{itemize}

        Answer: \{\{ Insert answer here \}\}
    \end{mdframed}
    \caption{\textbf{The prompt used for evaluating model's knowledge about forecasting questions}. 
    It asks the model to answer ``Yes'' or ``No'' given its pre-training knowledge and also allows for ``IDK''  (``I don't know''). See \autoref{sec:knowledge-eval} for the results.}
    \label{fig:model-knowledge-eval-prompt}
\end{figure}

%% file: tables/base_eval_zero_shot_by_category.tex
\begin{table}[ht!]
  \centering
  \resizebox{\textwidth}{!}{%
  \begin{tabular}{@{}lccccccccc@{}}
    \toprule
    \Large Model & \Large Arts \& Recreation & \Large Economics \& Business & \Large Education \& Research & \Large Environment \& Energy & \Large Healthcare \& Biology & \Large Politics \& Governance & \Large Science \& Tech & \Large Security \& Defense & \Large Sports \\
    \midrule 
    \Large GPT-3.5-Turbo & \Large 0.292 & \Large 0.281 & \Large 0.270 & \Large 0.245 & \Large 0.388 & \Large 0.244 & \Large 0.178 & \Large 0.235 & \Large 0.205 \\
    \Large GPT-3.5-Turbo-1106 & \Large 0.309 & \Large 0.294 & \Large 0.336 & \Large 0.239 & \Large 0.336 & \Large 0.343 & \Large 0.225 & \Large 0.250 & \Large 0.214 \\
    \Large GPT-4 (GPT-4-0613) & \Large 0.278 & \Large 0.260 & \Large 0.437 & \Large 0.201 & \Large 0.203 & \Large 0.228 & \Large 0.200 & \Large 0.224 & \Large 0.178 \\
    \Large GPT-4-1106-Preview & \Large 0.240 & \Large 0.244 & \Large 0.394 & \Large 0.222 & \Large 0.122 & \Large 0.218 & \Large 0.178 & \Large 0.207 & \Large 0.177 \\
    \Large Llama-2-7B & \Large 0.381 & \Large 0.356 & \Large 0.331 & \Large 0.359 & \Large 0.351 & \Large 0.399 & \Large 0.351 & \Large 0.288 & \Large 0.327 \\
    \Large Llama-2-13B & \Large 0.260 & \Large 0.247 & \Large 0.263 & \Large 0.218 & \Large 0.230 & \Large 0.245 & \Large 0.197 & \Large 0.222 & \Large 0.199 \\
    \Large Llama-2-70B & \Large 0.318 & \Large 0.329 & \Large 0.319 & \Large 0.299 & \Large 0.498 & \Large 0.329 & \Large 0.308 & \Large 0.264 & \Large 0.212 \\
    \Large Mistral-7B-Instruct & \Large 0.291 & \Large 0.265 & \Large 0.295 & \Large 0.228 & \Large 0.238 & \Large 0.271 & \Large 0.184 & \Large 0.236 & \Large 0.191 \\
    \Large Mistral-8x7B-Instruct & \Large 0.354 & \Large 0.272 & \Large 0.452 & \Large 0.256 & \Large 0.335 & \Large 0.252 & \Large 0.176 & \Large 0.227 & \Large 0.189 \\
    \Large Mixtral-8x7B-DPO & \Large 0.367 & \Large 0.315 & \Large 0.543 & \Large 0.213 & \Large 0.217 & \Large 0.287 & \Large 0.184 & \Large 0.265 & \Large 0.194 \\
    \Large YI-34B-Chat & \Large 0.263 & \Large 0.240 & \Large 0.332 & \Large 0.196 & \Large 0.208 & \Large 0.265 & \Large 0.196 & \Large 0.236 & \Large 0.212 \\
    \Large Claude-2 & \Large 0.293 & \Large 0.239 & \Large 0.326 & \Large 0.199 & \Large 0.226 & \Large 0.214 & \Large 0.175 & \Large 0.244 & \Large 0.194 \\
    \Large Claude-2.1 & \Large 0.293 & \Large 0.242 & \Large 0.316 & \Large 0.199 & \Large 0.226 & \Large 0.213 & \Large 0.183 & \Large 0.244 & \Large 0.194 \\
    \Large Gemini-Pro & \Large 0.301 & \Large 0.303 & \Large 0.432 & \Large 0.227 & \Large 0.210 & \Large 0.263 & \Large 0.175 & \Large 0.255 & \Large 0.189 \\
    \bottomrule
  \end{tabular}%
  }
  \caption{\textbf{Comparison of zero-shot Brier scores across categories and models} on the test set. This table lists the  Brier scores of all base models with respect to the specified categories.}
  \label{tab:zero-shot-cate}
\end{table}

%% file: tables/correlation_between_knowledge_and_zero_shot.tex
\begin{table}[ht!]
  \centering
  \begin{tabular}{lc}
    \toprule
    Category                & Score \\
    \midrule
    \textbf{Arts \& Recreation}      & \textbf{-0.417103} \\
    Economics \& Business   & -0.228040 \\
    \textbf{Education \& Research}   & \textbf{-0.359102} \\
    Environment \& Energy   & -0.135552 \\
    Healthcare \& Biology   & 0.162110 \\
    \textbf{Politics \& Governance}  & \textbf{-0.487266} \\
    Science \& Tech         & -0.091878 \\
    Security \& Defense     & -0.183253 \\
    Sports                  & -0.136017 \\
    \bottomrule
  \end{tabular}
    \caption{\textbf{Correlation between knowledge accuracy and zero-shot prompt Brier score by category}. Categories with an absolute correlation of 0.3 or greater, shown in bold, indicate a high correlation between accuracy on the training and validation set and forecasting performance  on the test set. This highlights that in certain domains model's forecasting capabilities are   correlated with its pre-training knowledge.}

   \label{tab:knowledge-zero-shot-correlation}
\end{table}

%% file: appendix/data.tex
\section{Dataset: Curation and Further Analysis}
\label{sec:data-details}

\subsection{Data Collection and Curation}

\paragraph{Scraping.}
To compile our dataset from the  forecasting platforms, we   query their  APIs or scrape the questions' webpages for initial data gathering.  For Metaculus, we first extract basic information via the API and  scrape the resolution criteria from webpage. INFER (CSET) and Good Judgment Open data are   gathered via web scraping, since no API provides the full data we need. Polymarket's data, except for   community predictions, is obtained from their API. Manifold's data is fully scraped via API. 
\input{figures/assign_category_prompt}
\input{figures/is_bad_title_prompt}

\paragraph{Assigning categories.}
There is no standard, uniform categorization of the forecast questions across the platforms.
We prompt GPT-3.5-Turbo to assign one of the 11 categories to each question. See \autoref{fig:categorization-prompt} for the category set and the  prompt we use.

\paragraph{Screening and curation.}
From manual examination, we notice that the initial dataset contains  questions that are  ambiguously formulated or overly personal. In a preliminary screening phase, we prompt GPT-3.5 to identify and exclude these   questions. See \autoref{fig:forecasting-question-assessment} for a prompt to detect   ill-defined questions, where we provide several few-shot examples. 

We then manually examine to eliminate all questions of low quality. This includes those with  few community forecasts or trading engagement on platforms such as Manifold and Polymarket, as well as any   ill-defined questions that GPT-3.5 is unable to identify during the initial screening.

\subsection{Further Statistics and Samples} 
We give a list of detailed statistics and plots on our data:
\begin{itemize}
    \item \autoref{fig:location-mentions} visualizes the location mentions in all the questions from our full dataset.
\item \autoref{tab:time-span-data}  gives the distribution of questions and forecasts across platforms in our full dataset.
    \item \autoref{tab: sample-data-point}  showcases a complete data sample in our curated set.
    \item \autoref{tab:list-sample-questions} shows  a list of questions with how community predictions shift over time.
    \item \autoref{fig:questions-by-quater}    shows the opening dates    of the questions in the full dataset.
    \item \autoref{fig:retrieval-retention} shows the percentage of questions that receives the retrieval  date at index $k= 1,2,3,4, 5$.
\end{itemize}
\input{figures/world_map}
\input{tables/stats_by_competition}
\input{tables/one_sample_question}
\input{tables/sample_questions}
\input{figures/resolve_stats}

%% file: figures/assign_category_prompt.tex
\begin{figure}[ht!]
    \centering
    \begin{mdframed}
        Question: \{question\}
        
        Background: \{background\}
        
        Options:
        \begin{itemize}
            \item Science \& Tech
            \item Healthcare \& Biology
            \item Economics \& Business
            \item Environment \& Energy
            \item Politics \& Governance
            \item Education \& Research
            \item Arts \& Recreation
            \item Security \& Defense
            \item Social Sciences
            \item Sports
            \item Other
        \end{itemize}
        
        Instruction: Assign a category for the given question.
        
        Rules:
        \begin{enumerate}
            \item Make sure you only return one of the options from the option list.
            \item Only output the category, and do not output any other words in your response.
            \item You have to pick a string from the above categories.
        \end{enumerate}
        
        Answer: \{\{ Insert answer here \}\}
    \end{mdframed}
    \caption{\textbf{Prompt for categorizing questions based on the provided options}. The prompt presents the forecasting question, along with 11 candidate category choices, and prompts the model to classify the question into one of the categories.}
    \label{fig:categorization-prompt}
\end{figure}

%% file: figures/is_bad_title_prompt.tex
\begin{figure}[ht!]
    \centering
    \begin{mdframed}
        I'm trying to assess the quality of an old forecasting dataset.

        Here is a forecasting question from the dataset: \{question\}.

        Please flag questions that don't sound like binary forecasting questions by outputting "flag". If it sounds like a reasonable question, output "ok". \\

        Examples of strings that should be flagged:
        \begin{itemize}
            \item Will I finish my homework tonight?
            \item Metaculus party 2023
            \item Will Hell freeze over?
            \item Heads or tails
            \item Will this video reach 100k views by the EOD?
        \end{itemize}

        Examples of strings that should not be flagged:
        \begin{itemize}
            \item Will Megan Markle and Prince Harry have a baby by the end of the year?
            \item Will the Brain Preservation Foundation's Large Mammal preservation prize be won by Feb 9th, 2017?
            \item Will there be more novel new drugs approved by the FDA in 2016 than in 2015?
        \end{itemize}

        If a question is already resolved, that doesn't mean it should be flagged. When in doubt, mark it as "ok".

        Your response should take the following structure:

        Insert thinking:
        \{\{ insert your concise thoughts here \}\}

        Classification:
        \{\{ insert "flag" or "ok" \}\}
    \end{mdframed}
    \caption{\textbf{The prompt for flagging ill-defined forecasting questions} in our dataset. The prompt contains several few-shot examples where the questions are ill-defined. A LM is prompted to filter out any questions of similar nature.}
    \label{fig:forecasting-question-assessment}
\end{figure}

%% file: figures/world_map.tex
\begin{figure}[ht]
  \centering
  \includegraphics[width=0.9\linewidth]{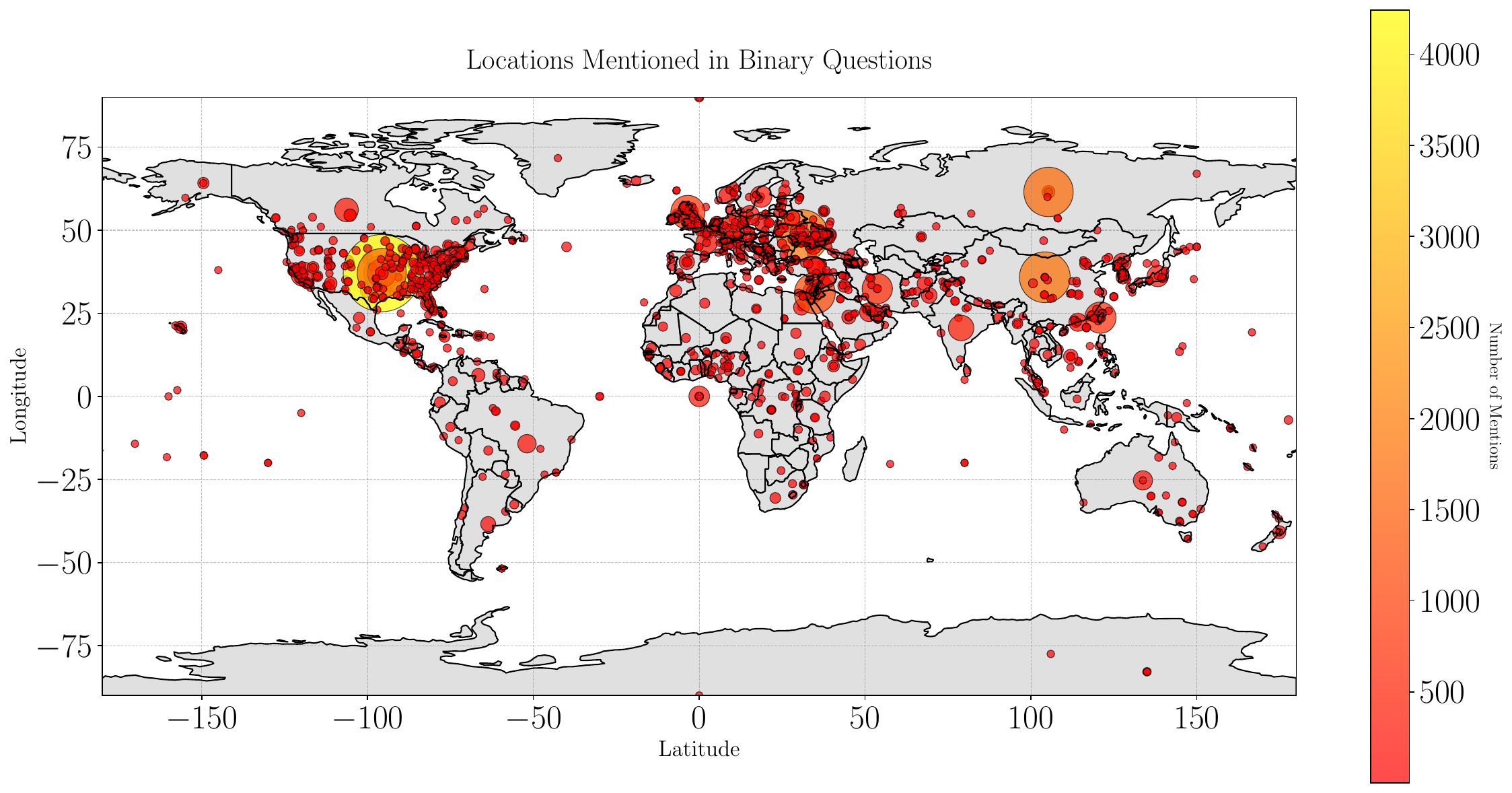}
  \caption{\textbf{Location mentions in all binary questions in our full dataset}. We visualize all location mentions in our full dataset on a world map. This shows that  the dataset provides a diverse coverage of topics across the globe.}
  \label{fig:location-mentions}
\end{figure}

%% file: tables/stats_by_competition.tex
\begin{table}[!ht]
  \centering
  \resizebox{\columnwidth}{!}{%
  \begin{tabular}{lcccccc}
    \toprule
    Platform & Questions (All) & Predictions (All) & Questions (Binary) & Predictions (Binary) & Brier Score (Binary) \\
    \midrule
    Metaculus    & $8,881$  & $638,590$   & $4,862$  & $387,488$  & $.130$ \\
    INFER & $308$    & $73,778$    & $192$    & $47,918$   & $.079$ \\
    GJOpen        & $2,592$  & $743,671$   & $1,168$  & $342,216$  & $.134$ \\
    Manifold      & $24,284$ & $1,997,928$ & $20,319$ & $1,387,668$ & $.155$ \\
    Polymarket    & $12,689$ & $3,720,640$ & $7,123$  & $1,879,035$ & $.158$ \\
    \bottomrule
\end{tabular}%
}
\caption{\textbf{Raw dataset statistics across platforms}. The Brier scores are calculated by averaging over all time points where the platforms provide crow aggregates.}
  \label{tab:time-span-data}
\end{table}

%% file: tables/one_sample_question.tex
\begin{table}[!htp]\centering
\scriptsize
\resizebox{\columnwidth}{!}{%
\begin{tabular}{lp{13cm}}\toprule 
\textbf{Field} & \textbf{Information} \\\midrule
\textbf{Question} & \textsc{Will Starship achieve liftoff before Monday, May 1st, 2023?} \\
\textbf{Start Date} & 2023-04-17 \\
\textbf{End Date} & 2023-04-30\\
\textbf{Resolve Date} & 2023-04-20\\
\textbf{Category} & Science \& Technology \\
\textbf{Platform} & Metaculus \\
\textbf{Resolution} & 1.0\\
\textbf{URL} & \url{https://www.metaculus.com/api2/questions/15973/} \\
\textbf{Background} & On April 14th, SpaceX received a launch license for its Starship spacecraft. A launch scheduled for April 17th was scrubbed due to a frozen valve. SpaceX CEO Elon Musk tweeted: ``Learned a lot today, now offloading propellant, retrying in a few days \ldots'' \\
      \textbf{Resolution Criteria} & This question resolves Yes if Starship leaves the launchpad intact and under its own power before 11:59pm ET on Sunday, April 30th. \\
\textbf{Community Predictions} & 
\begin{tabular}{@{}lp{6cm}@{}}
\multicolumn{2}{l}{(2023-04-17, 0.725), (2023-04-17, 0.793), (2023-04-17, 0.71), (2023-04-17, 0.704), (2023-04-17, 0.722),} \\
\multicolumn{2}{l}{(2023-04-17, 0.754), (2023-04-18, 0.74), (2023-04-18, 0.726), (2023-04-18, 0.707), (2023-04-18, 0.703),} \\
\multicolumn{2}{l}{(2023-04-18, 0.701), (2023-04-18, 0.698), (2023-04-18, 0.666), (2023-04-18, 0.665), (2023-04-18, 0.668),} \\
\multicolumn{2}{l}{(2023-04-18, 0.666), (2023-04-18, 0.63), (2023-04-18, 0.636), (2023-04-18, 0.652), (2023-04-18, 0.659),} \\
\multicolumn{2}{l}{(2023-04-18, 0.663), (2023-04-18, 0.664), (2023-04-18, 0.678), (2023-04-18, 0.687), (2023-04-18, 0.686),} \\
\multicolumn{2}{l}{(2023-04-18, 0.686), (2023-04-18, 0.686), (2023-04-18, 0.658), (2023-04-18, 0.658), (2023-04-18, 0.664),} \\
\multicolumn{2}{l}{(2023-04-18, 0.671), (2023-04-18, 0.677), (2023-04-18, 0.685), (2023-04-18, 0.685), (2023-04-18, 0.69),} \\
\multicolumn{2}{l}{(2023-04-18, 0.689), (2023-04-18, 0.691), (2023-04-18, 0.698), (2023-04-18, 0.706), (2023-04-18, 0.703),} \\
\multicolumn{2}{l}{(2023-04-18, 0.704), (2023-04-18, 0.706), (2023-04-18, 0.702), (2023-04-18, 0.703), (2023-04-18, 0.704),} \\
\multicolumn{2}{l}{(2023-04-18, 0.704), (2023-04-18, 0.702), (2023-04-18, 0.702), (2023-04-18, 0.702), (2023-04-18, 0.701),} \\
\multicolumn{2}{l}{(2023-04-18, 0.701), (2023-04-18, 0.689), (2023-04-18, 0.689), (2023-04-18, 0.686), (2023-04-18, 0.688),} \\
\multicolumn{2}{l}{(2023-04-18, 0.686), (2023-04-18, 0.684), (2023-04-19, 0.688), (2023-04-19, 0.688), (2023-04-19, 0.689),} \\
\multicolumn{2}{l}{(2023-04-19, 0.696), (2023-04-19, 0.695), (2023-04-19, 0.699), (2023-04-19, 0.697), (2023-04-19, 0.699),} \\
\multicolumn{2}{l}{(2023-04-19, 0.702), (2023-04-19, 0.703), (2023-04-19, 0.703), (2023-04-19, 0.705), (2023-04-19, 0.71),} \\
\multicolumn{2}{l}{(2023-04-19, 0.712), (2023-04-19, 0.713), (2023-04-19, 0.713), (2023-04-19, 0.714), (2023-04-19, 0.714),} \\
\multicolumn{2}{l}{(2023-04-19, 0.714), (2023-04-19, 0.717), (2023-04-19, 0.717), (2023-04-19, 0.713), (2023-04-19, 0.713),} \\
\multicolumn{2}{l}{(2023-04-19, 0.713), (2023-04-19, 0.717), (2023-04-19, 0.717), (2023-04-19, 0.716), (2023-04-19, 0.72),} \\
\multicolumn{2}{l}{(2023-04-19, 0.721), (2023-04-20, 0.721), (2023-04-20, 0.717), (2023-04-20, 0.716), (2023-04-20, 0.715),} \\
\multicolumn{2}{l}{(2023-04-20, 0.719), (2023-04-20, 0.723), (2023-04-20, 0.725), (2023-04-20, 0.725), (2023-04-20, 0.726),} \\
\multicolumn{2}{l}{(2023-04-20, 0.726), (2023-04-20, 0.73), (2023-04-20, 0.73), (2023-04-20, 0.728), (2023-04-20, 0.733),} \\
\multicolumn{2}{l}{(2023-04-20, 0.734)]} \\
\end{tabular}
\\
\textbf{Extracted URLs} & \url{https://www.youtube.com/live/-1wcilQ58hI}, \url{https://twitter.com/nextspaceflight/status/1648797064183128065},  \url{https://twitter.com/SciGuySpace/status/1648498635355865089}, \url{https://twitter.com/nextspaceflight/status/1648425030018293760}, \url{https://twitter.com/SpaceX/status/1648092752893313024} \\
\bottomrule
\end{tabular}%
}
\caption{\textbf{A sample question from our dataset with all its fields} (full version of \autoref{tab: short-data-point}). Each data point  consists of the following fields: question, start date, end date, resolve date, the final resolution, question category, platform, URL, background, resolution criteria, community predictions, and extracted URLs (from the background and comment section). The resolution is not presented to the model. We do not use the URLs that are extracted from the comment section, since certain comments may be made after the resolution.} \label{tab: sample-data-point}
\end{table}

%% file: tables/sample_questions.tex
\begin{table}[!ht]
  \centering
  \small 
  \renewcommand{\arraystretch}{1.5} 
  \setlength\tabcolsep{4pt} 
  \resizebox{\linewidth}{!}{%
    \begin{tabular}{p{7cm} p{2cm} p{1.6cm} p{1.6cm} p{1.6cm} p{1cm} p{1cm} p{1cm} r}
      \toprule
      \textbf{Sample Question} & \textbf{Category} & \textbf{Start Date} & \textbf{Close Date} & \textbf{Resolution Date} & \textbf{25\%} & \textbf{50\%} & \textbf{90\%} & \textbf{Answer} \\
      \midrule
      \textsc{Will AI doctors replace human doctors by the end of 2023?} & Science\hphantom{sss} \hphantom{xxx} \& Tech & 2023-07-27 & 2023-12-31 & 2023-12-30 & 0.073 & 0.087 & 0.102 & No \\
      \textsc{Will US CDC classify a SARS-CoV-2 variant as "high consequence" by August 1, 2022?} & Healthcare\hphantom{sss} \& Biology & 2021-07-31 & 2021-11-01 & 2022-08-02 & 0.39 & 0.408 & 0.412 & No \\
      \textsc{Will Coinbase file for bankruptcy in 2022?} & Economics\hphantom{sss} \& Business & 2022-05-12 & 2022-12-31 & 2023-01-01 & 0.08 & 0.079 & 0.072 & No \\
      \textsc{Will COP26 finalize the "Paris Rulebook" by November 16, 2021?} & Environment \& Energy & 2021-08-26 & 2021-11-13 & 2021-11-14 & 0.063 & 0.074 & 0.13 & Yes \\
      \textsc{Will Bongbong Marcos win the 2022 Philippine Presidential Election?} & Politics\hphantom{ssssss} \& Governance & 2021-12-20 & 2022-05-08 & 2022-05-26 & 0.759 & 0.752 & 0.759 & Yes \\
      \textsc{Will UC Berkeley be primarily in-person for Fall 2021?} & Education\hphantom{sss} \& Research & 2021-01-22 & 2021-08-01 & 2021-08-26 & 0.723 & 0.74 & 0.765 & Yes \\
      \textsc{Will Trump issue another NFT Collection before the 2024 Presidential Election?} & Arts\hphantom{sssssssss} \& Recreation & 2023-11-01 & 2023-12-12 & 2023-12-12 & 0.484 & 0.585 & 0.556 & Yes \\
      \textsc{Will a nuclear weapon be detonated in 2023 (including tests and accidents)?} & Security\hphantom{sssss} \& Defense & 2022-12-09 & 2023-12-31 & 2024-01-01 & 0.28 & 0.32 & 0.304 & No \\
      \textsc{Will Charlotte Hornets beat Detroit Pistons on Oct 27, 2023, in the NBA?} & Sports & 2023-10-16 & 2023-10-28 & 2023-10-28 & 0.46 & 0.513 & 0.337 & No \\
      \textsc{Will flight 1111 from Munich to Zurich on 2023-08-29 arrive on time or with more than 30 mins delay?} & Other & 2023-08-27 & 2023-08-29 & 2023-08-29 & 0.617 & 0.734 & 0.809 & Yes \\
      \bottomrule
    \end{tabular}%
  }
  \caption{\textbf{One sample question from each category} along with the community's predictions at different prediction dates (25\%, 50\%, and 90\% from the start date to resolve date). As the questions approach their resolution dates, the crowd's confidence in the outcome generally increases, reflecting the influence of new information.}
  \label{tab:list-sample-questions}
\end{table}

%% file: figures/resolve_stats.tex
\begin{figure}[ht]
    \centering
    \begin{subfigure}[b]{0.6\linewidth}
        \includegraphics[width=\linewidth]{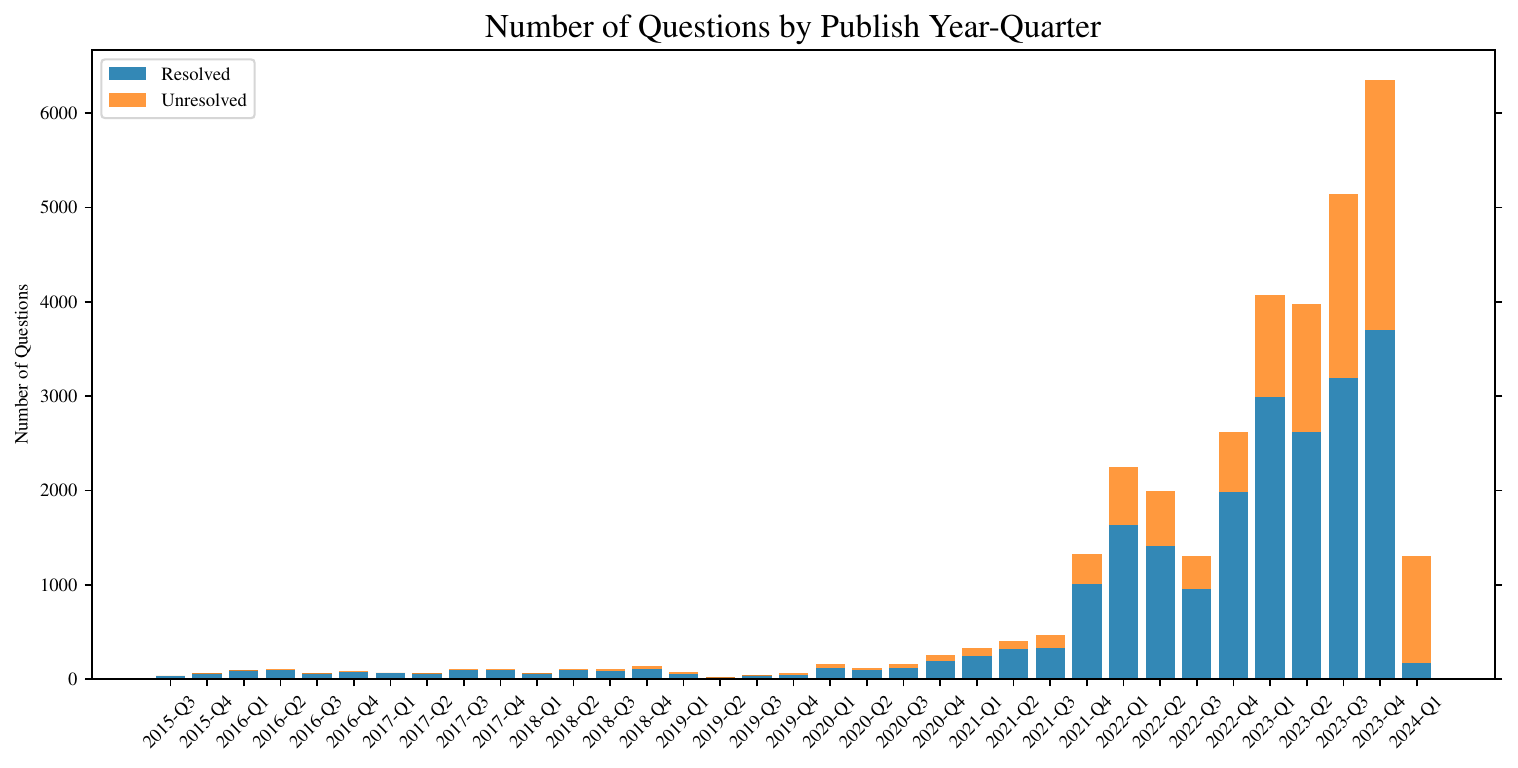}
        \caption{\textbf{Distribution of the opening dates of the questions in our full datasets}, ordered by year-quarter. Activity on these platforms has sharply increased over the past two years.}
        \label{fig:questions-by-quater}
    \end{subfigure}
    \hfill 
    \begin{subfigure}[b]{0.38\linewidth}
        \includegraphics[width=0.9\linewidth]{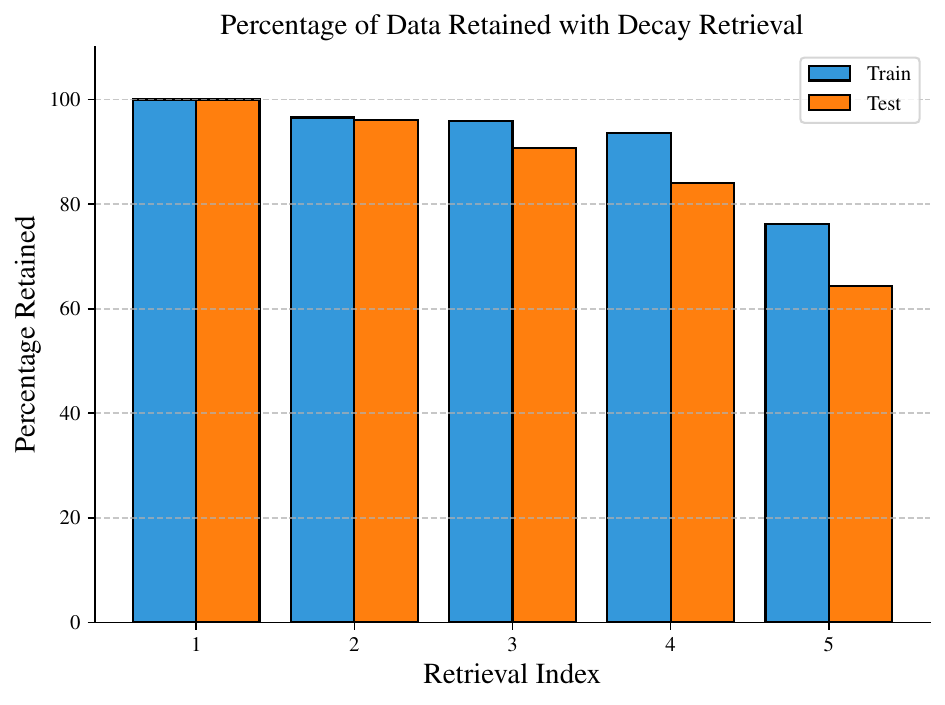} 
        \caption{\textbf{Percentage of questions that receive retrieval} at each retrieval index (1--5). The late retrieval indices can miss certain questions, since questions may resolve much earlier than the close time.}
        \label{fig:retrieval-retention}
    \end{subfigure}
    \caption{\textbf{Question publish time distribution and retrieval dates}.} 
\end{figure}

%% file: appendix/system.tex
\section{Details about Our System}
\label{sec:system-details}
We   provide details about our system,  described at a high-level in \autoref{sec:system}. 
 We specify the hyperparameters used in our (optimized) settings. Some of them are discovered by the hyperparameter sweep (\autoref{subsec:hp_sweep}).
\subsection{Retrieval System}
Our retrieval system consists of $4$ steps. We provide further details on each below.

    \paragraph{Step 1: Search query generation.} We identify two good   prompts to generate search queries in our hyperparameter sweep procedure, listed in \autoref{fig:both-sq-prompts}. Given a question, we ask GPT-4-Preview-1106 to generate $6$ search queries using both prompts (at $0$ temperature).    We take the union of all the resulting search queries along with the question itself to query the news API's. 

    \input{figures/search_query_prompt}
    \input{figures/summarization_prompt}
    
    \paragraph{Step 2: News retrieval.} On each news API and each search query, our system is set to retrieve the top  $10$ articles published within a given retrieval date range. We use the default ranking of each API and only retrieve English-language articles.  

    In cases where the background description of a question contains links to webpages, our system scrapes them, parses the clean texts, and presents the   summaries to the reasoning model. We take measures to  ensure that this leaks no information beyond the retrieval range.
   First,  we maintain a whitelist of news websites that publish timestamped articles and only retrieve from the whitelist. 
 Second,  our system  checks the publish date of each article  and discard it if the date is not available or outside the retrieval range. 
    
\input{figures/relevance_rating_prompt}
     \paragraph{Step 3: Relevance ranking.} We use GPT-3.5-Turbo to rate the relevance of an article with respect to a question at $0$ temperature. The prompt is given in \autoref{fig:relevance-prompt}. 
     
     Our retrieval system can retrieve a large number of texts (say, $>50$ articles) at the initial stage prior to relevance filtering.  To improve the run-time and save cost, we only present the article's title and  its first 250 words to the model in context for relevance rating.  In \autoref{sec:relevance}, we test that this well approximates the result from giving the full texts.

    The system rates the relevance of  each retrieved article at the scale of 1--6. 
    Any article that receives a rating of $\leq 3$ is discarded. We do not make an attempt to optimize this threshold or the prompt choice here.

     \paragraph{Step 4: Summarization.} We use GPT-3.5-Turbo to summarize the relevant articles. The temperature is set to be $0.2$. In cases where the article length exceeds the context window, we simply truncate it to fit the window size. We remark that our prompt (\autoref{fig:summarization-prompt}) also contains the question and its background description, and the model is instructed to keep any information in the article that is relevant to answering the question. \autoref{fig:summarization-prompt} shows the best prompt found via hyperparameter sweep on the validation set (\autoref{subsec:hp_sweep}).

\subsection{Reasoning System}
\input{figures/optimal_reasoning_prompt}
\begin{wrapfigure}{r}{0.45\textwidth}
    \centering
    \vspace{-3.2em}
    \fbox{
        \begin{minipage}{0.92\linewidth}
            {Question:}
            \{question\}\\[1ex]
            {Background:}
            \{background\}\\[1ex]
            {Resolution criteria:}
            \{criteria\}\\[1ex]
            {Today's date:} \{date\_retrieval\}\\
            {Question close date:} \{date\_end\}\\[1ex]
            We have retrieved the following information:\\
            \{retrieved\_info\}
        \end{minipage}
    }
    \caption{\textbf{All scratchpad prompts  begin with a  question's basic information, followed by retrieval.} The fine-tuned model only takes this information and requires no further instructions.}
    \vspace{-1.6em}
    \label{fig:basic-prompt-reasoning}
\end{wrapfigure}
We use both GPT-4-1106-Preview and our fine-tuned GPT-4 to generate forecasts. 
We prompt the former with our top $3$ reasoning prompts, including \autoref{fig:optimal-reasoning-prompt}. The other prompts also conform to the basic template as shown in \autoref{fig:basic-prompt-reasoning}, though with different scratchpad reasoning instructions following the retrieved information section.
The fine-tuned model does not require detailed scratchpad instructions (\autoref{subsec:fine_tuning}). Thus, \autoref{fig:basic-prompt-reasoning} is the entire prompt structure   to the fine-tuned model to elicit its reasonings.

In addition, as we remarked in \autoref{subsec:fine_tuning}, Claude-2.1 was prompted to generate reasoning-prediction pairs for fine-tuning. However, it is not directly used  for reasoning in our system.

%% file: figures/search_query_prompt.tex
  \begin{figure}[tbp]
  \vspace{-9em}
\centering
\begin{subfigure}[b]{0.98\textwidth}
     \begin{mdframed}
I will provide you with a forecasting question and the background information for the question.\\

        Question: \{question\}
        
        Background: \{background\}
        \\

    Task:\\
- Generate brief search queries (up to \{max\_words\} words each) to gather information on Google that could influence the forecast.\\

You must generate this exact amount of queries: \{num\_keywords\}
\\

Your response should take the following structure:
Thoughts:
\{\{ Insert your thinking here. \}\}
\\
Search Queries:
\{\{ Insert the queries here. Use semicolons to separate the queries. \}\}
    \end{mdframed}
    \caption{A straighforward search query expansion prompt}
\label{fig:sq-prompt-2}
\end{subfigure}
\vspace{1.0em}
\begin{subfigure}[b]{0.98\textwidth}
    \begin{mdframed}
I will provide you with a forecasting question and the background information for the question. I will then ask you to generate short search queries (up to \{max\_words\} words each) that I'll use to find articles on Google News to help answer the question.    \\
    
        Question: \{question\}
        
        Background: \{background\}
        \\
    
You must generate this exact amount of queries: \{num\_keywords\}
\\

Start off by writing down sub-questions. Then use your sub-questions to help steer the search queries you produce.
\\

Your response should take the following structure:
Thoughts:
\{\{ Insert your thinking here. \}\}
\\
Search Queries:
\{\{ Insert the queries here. Use semicolons to separate the queries. \}\}
    \end{mdframed}
    \caption{The second search query prompt we use. It first asks the model to consider sub-questions and use that to steer the outputs.}
    \label{fig:sq_prompt}
\end{subfigure}
\vspace{-1.8em}
\caption{\textbf{Prompts to generate search queries} based on the question's data. }
\label{fig:both-sq-prompts}
\end{figure}

%% file: figures/summarization_prompt.tex
\begin{figure}[ht!]
    \centering
    \vspace{-4em}
\begin{mdframed}
I want to make the following article shorter (condense it to no more than 100 words).\\

Article:
\{article\}

When doing this task for me, please do not remove any details that would be helpful for making considerations about the following forecasting question. \\

Forecasting Question: \{question\}

Question Background: \{background\}
\end{mdframed}
 \caption{\textbf{The summarization prompt we use in our retrieval system}. The prompt provides a question, its background, and a relevant article. It asks the LM to condense the article without removing any information relevant to the forecasting question.}
    \label{fig:summarization-prompt}
\end{figure}

%% file: figures/relevance_rating_prompt.tex
\begin{figure}[ht!]
    \centering
\begin{mdframed}
Please consider the following forecasting question and its background information.
After that, I will give you a news article and ask you to rate its relevance with respect to the forecasting question.
\\

Question:
\{question\}

Question Background:\{background\}

Resolution Criteria:\{resolution\_criteria\}
\\

Article:
\{article\}
\\

Please rate the relevance of the article to the question, at the scale of 1-6

1 -- irrelevant

2 -- slightly relevant

3 -- somewhat relevant

4 -- relevant

5 -- highly relevant

6 -- most relevant
\\ 

Guidelines:

- You don't need to access any external sources. Just consider the information provided.

- Focus on the content of the article, not the title.

- If the text content is an error message about JavaScript, paywall, cookies or other technical issues, output a score of 1.
\\

Your response should look like the following:

Thought: \{\{ Insert your thinking  \}\}

Rating: \{\{ Insert answer here \}\}
\end{mdframed}
 \caption{\textbf{Prompt used to rate the relevance of an article with respect to a question}. The prompt asks a LM to rate the relevance of an article with respect to a question at the scale of 1--6. We extract the numerical value following ``Rating:''.}
    \label{fig:relevance-prompt}
\end{figure}

%% file: figures/optimal_reasoning_prompt.tex
\begin{figure}[ht!]
    \centering
\begin{mdframed}
Question: \{question\} \\

Question Background: 
\{background\} \\

Resolution Criteria: 
\{resolution\_criteria\} \\

Today's date: \{date\_begin\} \\
Question close date: \{date\_end\} \\

We have retrieved the following information for this question: \\
\{retrieved\_info\} \\

Instructions: \\
1. Given the above question, rephrase and expand it to help you do better answering. Maintain all information in the original question. \\
\{\{ Insert rephrased and expanded question. \}\} \\

2. Using your knowledge of the world and topic, as well as the information provided, provide a few reasons why the answer might be no. Rate the strength of each reason. \\
\{\{ Insert your thoughts \}\} \\

3. Using your knowledge of the world and topic, as well as the information provided, provide a few reasons why the answer might be yes. Rate the strength of each reason. \\
\{\{ Insert your thoughts \}\} \\

4. Aggregate your considerations. Think like a superforecaster (e.g. Nate Silver). \\
\{\{ Insert your aggregated considerations \}\} \\

5. Output an initial probability (prediction) given steps 1-4. \\
\{\{ Insert initial probability \}\} \\

6. Evaluate whether your calculated probability is excessively confident or not confident enough. Also, consider anything else that might affect the forecast that you did not before consider (e.g. base rate of the event). \\
\{\{ Insert your thoughts \}\} \\

7. Output your final prediction (a number between 0 and 1) with an asterisk at the beginning and end of the decimal. \\
\{\{ Insert your answer \}\} \\
\end{mdframed}
 \caption{\textbf{The scratchpad reasoning prompt that gets lowest Brier score on validation set}. The prompt first provides the basic information about the question, along with retrieved article summaries. Then it gives   instructions to guide the model's reasoning path (\autoref{sec:system}).}
    \label{fig:optimal-reasoning-prompt}
\end{figure}

%% file: appendix/opt.tex
\section{Details on Optimization of Our System}
\label{sec:appendix-opt}
\subsection{Hyperparameter Sweep}
\label{sec:appendix-hp-sweep}
Throughout the experiment, we set the retrieval date to be the midpoint between a question's open and resolve date. At this time point, the crowd aggregates achieve $0.160$ Brier score, averaged over all questions in our validation set. 

All the hyperparameter sweeps below evaluate all questions in the validation set.

 \paragraph{Search query prompt.} We sweep over $6$ candidate prompts for generating search queries. The top $2$ prompts lead to retrieved articles that have average relevance rating of $3.08$ and $3.09$, while other prompts below $3.04$. Among all articles with rating at least $4$, the average rating is $4.37$ and $4.38$ via the top $2$ search query prompts, which is  also the highest among all candidates.

\paragraph{Summarization prompt.} We sweep over $5$ candidate prompts for summarization and evaluate the resulting Brier scores. 
The best summarization prompt gives a Brier score of $0.193$ and the second gives $0.201$. In this step of hyperparameter search,   the ordering of the summaries,   article count and reasoning prompt are randomly chosen for each question to avoid confounding.

\paragraph{Article ordering and count.}
In this step, we sweep over both   orderings of articles (by recency or relevance), and  over   $5$ candidate choices of   $k\in[5,10,15,20,30]$. We run our full system on all questions in the validation set. 
Presenting $15$ article summaries and ordering them by relevance gives the lowest Brier score of $0.177$ on GPT-4-1106-Preview.   Similar performance can be achieved by presenting $20$ articles summaries.

\paragraph{Reasoning prompt.} 
We hand-craft  $15$ prompts for eliciting forecasts and reasonings. 
The best prompt (\autoref{fig:optimal-reasoning-prompt}) achieves  
a Brier score of $0.167$ on the validation set (while fixing  the optimal hyperparameter choices found by the optimization   stages above). Two other top prompts get  $0.170$ and $0.174$.   The  best prompt is given in \autoref{fig:optimal-reasoning-prompt}.

\input{tables/ensemble_eval}
\paragraph{Ensembling.} We implement $5$ ensembling methods, including mean, median, geometric mean, trimmed mean,  and a variant of universal self-consistency (USC) \citep{chen2023universal}. The last two approaches are defined as follows:
\begin{itemize}
    \item For  the trimmed mean, we assign uniform weights over the    input   forecasts, reduce the weight of the forecast furthest from the median  by half, redistribute the  half weight uniformly to the other forecasts, and finally output the weighted average. We remark that this is not a standard implementation of  trimmed mean, and it is set this way since we only aggregate a small number (i.e., 6) of forecasts in our system. 
    \item For USC, we  present the (external) reasoning-prediction pairs  to a final LM, which is then  prompted to form an aggregated forecast. In this hyperparameter sweep, we use GPT-4-1106-Preview as the aggregator model.
\end{itemize}
We evaluate all these methods on the validation set by generating $6$ base reasonings for ensembling, using our optimal system setup. Trimmed mean   achieves the lowest Brier score; see   \autoref{tab:brier_scores-ensemble} for the results.  The USC method, in contrast, does not demonstrate improvement over the baseline.

\subsection{News API Evaluations}
\label{sec:news-api-eval}
We   justify our choice of the news API's. To begin with, we experiment with $5$ eligible   APIs to news corpus that accept retrieval date ranges, which, for our purpose, must be  specified   to prevent leakage.  In particular, we assess Google News (accessed via Python open source package \texttt{\small gnews}), NewsCatcher, Newsdata.io, Aylien,  and  NewsAPI.org.

 To assess the quality of their retrievals,  we first take $24$ unresolved forecasting questions. Next, we  prompt  GPT-4-1106-Preview to generate two search queries for each of these 24 questions, similar to the first stage of our retrieval system (\autoref{sec:system}). We   use these queries to search for articles via all $5$ APIs, 
  restricting the retrieval range to the last 24 hours.
  
  Finally,  we prompt GPT-4 to rate the relevance of the articles with respect to the original questions. Higher scores indicate greater relevance. We compute the sum of scores of all retrieved articles for each API option. As a result, NewsCatcher and Google News achieve the highest scores of 35 and 39, respectively. The other three APIs,  Newsdata.io, Aylien, and NewsAPI.org, score 16.5, 30.5, and 23.5.

\subsection{Relevance Rating Approximations}
\label{sec:relevance}
We prompt  GPT-3.5 Turbo to score the relevance of all retrieved articles (from Google News and NewsCatcher) with respect to the question.   Our prompt is given by \autoref{fig:relevance-prompt}, where the question's metadata along with an article text is provided to the model in context. 
The prompt asks the LM to rate the relevance of  an article to the given question at the scale of 1--6, where 6 is considered ``most relevant'' and 1 ``irrelevant''.
In our system, we filter out articles with ratings below $4$. 

\paragraph{Methods.}
Due to cost constraints, we cannot afford to evaluate  relevance scores using the full article texts. We experimentally explore $3$ cost-saving alternatives    to approximate full text-based ratings:
\begin{enumerate}[(i)]
    \item Title-only. We only give the article title to the model for relevance rating. Unfortunately, via manual inspection, we find the  web scraper\footnote{\href{https://pypi.org/project/newspaper4k/}{https://pypi.org/project/newspaper4k/}}  sometimes fails to to load the full text of an article page, usually due to   rendering error  or paywall. In such cases, the article content may be incomplete or simply a error message, whereas the title is retrieved and appears relevant. Consequently, the model may be misled by the title. Therefore, we believe that this approach  is unviable.
    \item Title + first 250 words. We give the article title and the first 250 words to the model for rating.
    \item Embedding similarity. We embed the article text and question text metadata and compute their cosine similarities.  We threshold by the similarities to filter articles. 
\end{enumerate}

\paragraph{Experiment.}
We experiment with approach (ii) and (iii) above. For (ii), we prompt GPT-3.5-Turbo and Mixtral-8x7B-DPO for relevance rating with the same prompt template (\autoref{fig:relevance-prompt}). For (iii), we use OpenAI's \texttt{\small text-embedding-3-large} as the embedding model.

To generate raw articles, we randomly sample $15$ questions from the validation set, run our retrieval system up to relevant rating, and collect $169$ articles in total.

We evaluate   the  relevance scores of the full texts via GPT-4 (considered as gold labels) and compare with the approximations described above. An article is said be relevant if its rating is at least $4$ from the full text query via GPT-4. We compute the recall and precision of the approximate methods as follows. 

\begin{itemize}
    \item At a threshold of $4$,  title + first $250$ words via GPT-3.5-Turbo gives   recall of $0.73$ and precision of $0.65$. 
        \item At a threshold of $3$,  title + first $250$ words via Mixtral-8x7B-DPO gives   recall of $0.70$ and precision of $0.63$. 
    \item At a threshold of $0.48$, the cosine similarity   method gives recall of $0.73$ and precision of $0.54$. 
\end{itemize}
In sum, fixing recall at above $70\%$, querying  GPT-3.5-Turbo with title + first $250$ words attains the highest precision. The average token length of the articles is $1087.6$. Hence, the method of using first $250$ words (or roughly $330$ tokens) saves about $70\%$ of the cost. 



%% file: tables/ensemble_eval.tex
\begin{wraptable}{r}{8cm} 
\vspace{-1em}
  \centering
  \begin{tabular}{lr}
    \toprule
    Ensemble Method & Brier Score \\
    \midrule
    Mean & 0.1656 \\
    {Median} & {0.1651} \\
    Geometric Mean & 0.1655 \\
    \textbf{Trimmed Mean} & \textbf{0.1649} \\
    USC \citep{chen2023universal}  & 0.1691 \\
    \midrule
    Baseline (No Ensemble) & 0.1676 \\
    \midrule
   Human  Crowd  & 0.1600 \\
    \bottomrule
  \end{tabular}
  \caption{\textbf{Brier scores across different ensembling methods on the validation set.}   ``Baseline'' refers to the average Brier score of the base predictions (i.e., the inputs to ensembling).}
  \label{tab:brier_scores-ensemble}
  \vspace{-2em}
\end{wraptable}

%% file: appendix/examples.tex
\section{Example Outputs}
\label{sec:examples}
We provide $5$ qualitative example outputs  from our fine-tuned GPT-4 model on the test set.  
\begin{figure}[h!]
\vspace{-0em}
 \begin{mdframed}
     1. Rephrased and Expanded Question:
Is it likely that Kevin McCarthy, the former House Speaker, will make a public announcement or have his resignation
reported by credible sources, indicating his departure from the United States House of Representatives, before the 
end of November 30, 2023? This question takes into account the current political climate, McCarthy's recent loss of
the Speaker position, his consideration of not seeking re-election, and the potential impact of his decision on the
Republican Party and the 20th District of California. The resolution of this market hinges on whether McCarthy will
officially declare his intention to vacate his seat within the specified timeframe.

2. Reasons why the answer might be no:

- McCarthy has not made a final decision and has expressed uncertainty about his future in Congress, which could 
mean he is still weighing his options and may not choose to resign within the given timeframe. (Strength: Strong)

- The articles suggest that while there is speculation and consideration of resignation, there is no definitive 
statement or action indicating an imminent resignation. (Strength: Strong)

- McCarthy may be waiting to see how the political landscape shifts or to make a decision after discussing with his
family, which could extend beyond the November 30 deadline. (Strength: Moderate)

- Given his history of strong financial support and political influence, McCarthy may feel compelled to stay and 
influence the political process, especially with the upcoming 2024 elections. (Strength: Moderate)

3. Reasons why the answer might be yes:

- There are reports and hints from McCarthy himself that he is considering resignation and may want to "get the 
hell out," which could indicate a potential for an early departure. (Strength: Moderate)

- The pressure from political opponents and the recent loss of the Speaker position may contribute to a decision to
resign, especially if he feels his influence is waning. (Strength: Moderate)

- The possibility of a special election being called could provide an opportunity for a controlled transition of 
power, which might be appealing to McCarthy. (Strength: Weak)

4. Aggregate Considerations:

While there is speculation and some suggestive comments from McCarthy about the possibility of resignation, there 
is no concrete evidence or announcement that he will resign by the specified date. His history of fundraising and 
political influence, as well as the lack of a definitive statement about his future, suggest that a resignation by 
November 30 is not certain.

5. Initial Probability:

Given the considerations above, the initial probability of McCarthy resigning by November 30 seems low.

6. Adjusted Probability Considerations:

- The calculated probability should not be excessively confident in either direction due to the uncertainty 
expressed by McCarthy himself.

- The base rate for such high-profile resignations in the face of uncertainty is not well-established, which adds 
to the difficulty in predicting this event.

- Intuitively, politicians often weigh their options carefully, and a decision to resign is not typically rushed, 
especially without a clear strategic reason.

7. Final Answer: *0.20*
 \end{mdframed}
 \caption{A response of the fine-tuned model on the question \href{https://polymarket.com/event/will-kevin-mccarthy-resign-from-the-house-by-nov-30?tid=1708585271779}{``Will Kevin McCarthy resign from the House by Nov 30?''} (resolved NO) at the last retrieval index. Inspecting the original prompt (not shown here), we notice that our retrieval system presents articles, published on November 30, that clearly state  that no  official announcement had yet been made about Kevin McCarthy's potential resignation. On the day of November, 30, the community prediction had dropped to $1\%$, but  our system hedges by claiming $20\%$.}
    \label{fig:example-kevin}
\end{figure}

\begin{figure}
\vspace{-3em}
\begin{mdframed}
1. Will Reddit make any official announcements prior to the 1st of July, 2023, regarding alterations to the pricing
structure or a postponement of the implementation of fees for its Application Programming Interface (API), which is
currently set to charge developers for higher usage? This question will be resolved affirmatively if Reddit 
communicates any of the following changes: a different pricing model than the current "\$0.24 per 1K API calls," 
adjustments to the free API usage thresholds, a delay in the implementation date beyond July 1, 2023, or any 
negotiations for specific rates with individual organizations do not count. The resolution of this question will be
determined by Metaculus, and the question may be annulled if the announcement is unclear or does not meet the 
specified criteria.

2. Reasons why the answer might be no:

- Reddit's CEO, Steve Huffman, has consistently shown a firm stance on the API changes, emphasizing the necessity 
for the company to move forward with the new pricing structure to ensure profitability and sustainability. This 
resolute position suggests that a change in the announced API fee pricing is unlikely.

- Despite the significant backlash and coordinated protests from thousands of Reddit communities, there has been no
indication from Reddit's internal communications or public statements that they are considering altering the API 
fee pricing or its implementation timeline.

- The internal memo from Reddit's CEO, where he mentioned that the protest "will pass" and that there has been no 
significant revenue impact from the protests, indicates that the company may be prepared to weather the discontent 
and proceed with the changes as planned.

3. Reasons why the answer might be yes:

- The widespread protest and the large number of communities going dark could potentially lead to a significant 
user and developer exodus, which might force Reddit to reconsider its position to avoid a substantial negative 
impact on its platform's activity and content.

- The specific mention of negotiations with individual organizations for different rates could be a sign that 
Reddit is open to adjusting the pricing for certain parties, which might lead to broader changes if the protests 
continue or if key developers threaten to shut down their apps.

- The negative publicity and user dissatisfaction could reach a tipping point where the company decides that the 
long-term damage to its brand and community is not worth the short-term financial gain, prompting a change in the 
API fee pricing policy.

4. Rating the strength of each reason:

- The strength of the resolute stance of Reddit's CEO is high, as it is based on consistent public statements and 
the company's broader financial strategy.

- The lack of any indication of a potential change in the API fee pricing is also a strong reason, given the 
proximity to the implementation date and the current lack of communication suggesting otherwise.

- The internal memo's dismissive tone about the protests and lack of revenue impact is a moderate reason, as it 
reflects the current mindset of the company but does not preclude a change in response to escalating pressure.

- The potential for a user and developer exodus is a moderate reason, as it is speculative and depends on the 
actual impact of the protests, which has not been fully realized yet.

- The mention of negotiations with individual organizations is a weak reason for a broader change, as it does not 
necessarily indicate a willingness to change the policy for all developers.

- The possibility of a tipping point due to negative publicity is a moderate reason, as it is based on the 
assumption that Reddit will prioritize its long-term brand and community over short-term financial gains, which is 
uncertain.

5. Aggregate your considerations:

Considering the strong and consistent stance of Reddit's leadership, the lack of any public indications of a 
change, and the dismissal of the protests' impact on revenue, it seems unlikely that Reddit will announce changes 
to its API fee pricing before July 1, 2023. However, the potential for negative publicity and the impact of the 
protests on the platform's activity cannot be entirely dismissed, which introduces some uncertainty into the 
forecast.

6. Output your answer:
Given the information available and the considerations above, the likelihood of Reddit announcing changes or a 
delay to its proposed API fee pricing before July 1, 2023, seems low. My answer, therefore, is: *0.2*
\end{mdframed}
\caption{A response from our fine-tuned model on the question \href{https://www.metaculus.com/questions/17469/reddit-api-pricing-change-before-july-1/}{``Will Reddit announce changes or a delay to its proposed API fee pricing before July 1, 2023?''} (resolved NO). The forecast was made at the retrieval date June, 14, 2023, one day after the question was opened. At the time, the community gives a forecast of $40\%$, where this response suggests $20\%$ which turned out to be closer to the resolution.}
\end{figure}

\begin{figure}[htbp]
    \centering
    \vspace{-2em}
\begin{mdframed}
1. Rephrased and Expanded Question:

Considering the current geopolitical tensions and recent historical events, is it likely that the State of Israel 
will engage in a military operation that results in the death of at least five Iranian nationals within Iran's 
borders, and subsequently either publicly claim responsibility for this action or execute the operation in such a 
clear and overt manner that credible sources can unambiguously attribute the attack to Israel, all before the end 
of the year 2023?

2. Decision Tree Outline:

- Yes Outcome:

 \quad  - Israel conducts a military operation against Iran.
  
   \quad  \quad - The operation is overt and clearly attributable to Israel.
    
    \quad \quad - Israel publicly acknowledges and takes credit for the operation.
    
  \quad - The operation results in at least five Iranian deaths within a 24-hour period.
  
- No Outcome:

 \quad  - Israel does not conduct a military operation against Iran.
 
  \quad - Israel conducts an operation, but it is covert and not attributable.
  
  \quad - Israel conducts an operation, but does not publicly acknowledge it.
  
 \quad  - The operation does not result in at least five Iranian deaths within a 24-hour period.

3. Branch Probability Analysis:

- Yes Outcome:

 \quad  - Given the recent escalation in tensions and historical precedents, there is a non-negligible chance that Israel
could conduct a military operation against Iran.

  \quad - The likelihood of Israel publicly acknowledging such an operation is lower, as they often maintain plausible 
deniability.

  \quad - The probability of an operation resulting in at least five Iranian deaths is uncertain and would depend on the 
scale and nature of the operation.

- No Outcome:

 \quad  - The probability of no military operation or a covert operation is higher, given Israel's historical preference 
for covert actions and the potential for international backlash.

  \quad - The likelihood of Israel not acknowledging an operation is also high, as this is a common practice to avoid 
direct conflict and international pressure.

  \quad - The chance of an operation not meeting the death threshold is also possible, especially if the operation is 
designed to be limited or surgical.

4. Discussion on Wildcards:

- A significant escalation in the conflict between Israel and Iran, such as a direct attack on Israeli soil 
attributed to Iran, could increase the probability of a 'Yes' outcome.

- International diplomatic efforts or significant pressure on Israel to de-escalate could decrease the probability 
of a 'Yes' outcome.

- Unpredictable events, such as a major terrorist attack or a sudden change in leadership in either country, could 
also impact the forecast.

5. Initial Probability (Prediction):

Given the current information and the outlined decision tree, the initial probability of a 'Yes' outcome seems low 
but not impossible. I would estimate it to be around 0.20.

6. Adjusted Probability Thoughts:

- The calculated probability may be too confident given the volatile nature of the Middle East and the 
unpredictability of military engagements.

- The recent death of an Iranian general in an Israeli airstrike could escalate tensions, potentially increasing 
the likelihood of a 'Yes' outcome.

- The lack of direct evidence linking Iran to the October 7 Hamas attack and the potential for covert operations 
could decrease the likelihood of a 'Yes' outcome.

7. Final Answer:
*0.15*
\end{mdframed}
    \caption{A response from our fine-tuned model on the question \href{https://www.metaculus.com/questions/19332/israel-deadly-attack-on-iran-before-2024/}{``Will Israel carry out and explicitly acknowledge a deadly attack on Iran before 2024?''} (resolved NO). In our fine-tuning phase, one of the prompts we use to generate strong reasonings asks the model to build a decision tree and assign probabilities.  The fine-tuned model learns the reasoning path (without being explicitly prompted to do so).}
    \label{fig:example-1}
\end{figure}

\begin{figure}
\begin{mdframed}
    1. Rephrased and Expanded Question:
    
Is it confirmed that former President Donald J. Trump will be present and actively participate in the inaugural 
Republican National Committee (RNC) presidential primary debate for the 2024 election cycle, which is scheduled to 
take place on August 23, 2023? The criteria for his participation include him speaking at least once during the 
debate and being recognized by the debate's moderator. This applies even if he joins the debate remotely. The 
resolution of this question will consider all segments of the debate if it is spread over multiple days. However, 
if the debate is postponed beyond September 30, 2023, the market will resolve to "No." The resolution will be based
primarily on the actual debate footage, supplemented by a consensus from credible news reports.

2. Reasons why the answer might be no:
- Trump may choose to skip the debate to maintain a strategic advantage or to avoid early confrontations with other
candidates.

- He might not want to share the stage with candidates he perceives as less significant, which could dilute his 
brand or message.

- There could be concerns about the debate format or the choice of moderators, which might lead to a decision not 
to participate.

3. Reasons why the answer might be yes:

- Participating in the debate would give Trump a platform to outline his policies and vision, which could be 
beneficial for his campaign.

- It would allow him to directly engage with and challenge his opponents, potentially strengthening his position.

- Trump might want to demonstrate leadership and confidence by being present at the first major event of the 
primary season.

4. Rating the strength of each reason:

- The strategic advantage of skipping the debate: Moderate. Trump has a history of unconventional campaign 
strategies, but skipping the first debate is a risky move.

- Avoiding sharing the stage: Moderate. Trump's brand is strong, but he may not want to risk being overshadowed or 
challenged by multiple opponents.

- Concerns about format or moderators: Moderate to High. Trump has previously expressed dissatisfaction with debate
arrangements, which could influence his decision.

- Benefiting from outlining policies: High. The debate is a significant opportunity to reach a large audience and 
make a case for his candidacy.

- Engaging with opponents: High. Trump is known for his confrontational style, which could play out to his 
advantage in a debate setting.

- Demonstrating leadership: High. As a former president, Trump might want to assert his dominance in the Republican
field from the outset.

5. Aggregate your considerations:

Considering the reasons and their strengths, it seems there is a balanced set of factors that could influence 
Trump's decision to participate or not. His history of unconventional tactics and concerns about debate 
arrangements could lead to a "No," while the potential benefits of participating in terms of policy promotion, 
engaging opponents, and demonstrating leadership suggest a "Yes." The decision will likely depend on the specific 
circumstances closer to the debate date and Trump's assessment of the strategic value of his participation.

6. Output your answer:
Given the balanced considerations and the unpredictable nature of Trump's campaign strategies, I would assign a 
probability that leans slightly towards participation due to the high benefits it offers, but not overwhelmingly so
due to the potential strategic considerations against it.

*0.55*
\end{mdframed}
    \caption{A response from our fine-tuned model on the question \href{https://polymarket.com/event/will-trump-attend-the-first-rnc-debate?tid=1708586828523}{``Will Trump attend the first RNC debate?''} (resolved NO on August, 23, 2023). The forecast was made at the retrieval index $1$ on June 15, 2023, $2$ days after the question was opened. At the time, the community predicted $60\%$, close to our model's forecast of $55\%$ from this response.}
    \label{fig:example-trump-rnc}
\end{figure}

\begin{figure}[htbp]
\begin{mdframed}
1. Rephrased and Expanded Question:

Is it accurate to state that the film 'Barbie' will achieve a domestic box office gross that is at least double the
amount that 'Oppenheimer' will earn during their respective opening weekends? This comparison will be based on the 
final reported figures for the three-day opening weekend (July 21-23) in the United States, as documented on Box 
Office Mojo. The outcome will be determined by comparing the "Domestic Weekend" grosses of 'Barbie' and 
'Oppenheimer,' and the market will resolve in favor of 'Yes' if 'Barbie's gross is at least twice that of 
'Oppenheimer.' If no final data is available by July 31, 2023, an alternative credible source will be used to 
resolve the market.

2. Reasons why the answer might be no:

- 'Oppenheimer' could have a stronger than expected performance, reducing the ratio between the two films' grosses.
(Strength: Weak, given the current projections and reports.)

- There could be a significant underestimation of 'Oppenheimer's appeal to a specific demographic that turns out in
large numbers. (Strength: Weak, as current data does not support this.)

- Unforeseen events could negatively impact 'Barbie's attendance, such as controversies or widespread technical 
issues in theaters. (Strength: Weak, as there is no indication of such events.)

3. Reasons why the answer might be yes:

- 'Barbie' has already been reported to have a significantly higher opening weekend gross than 'Oppenheimer,' with 
figures exceeding the 2x mark. (Strength: Strong, based on multiple reports.)

- 'Barbie' is playing in more theaters than 'Oppenheimer,' which could contribute to a higher gross. (Strength: 
Moderate, as screen count can influence box office performance.)

- 'Barbie' has broader appeal and targets a wider demographic, including families and younger audiences, which 
could lead to higher attendance. (Strength: Moderate, as demographic appeal is a significant factor in box office 
performance.)

4. Aggregate your considerations:

Considering the information provided, 'Barbie' has a strong lead over 'Oppenheimer' in terms of opening weekend 
gross, with multiple reports indicating that it has already achieved more than double the gross of 'Oppenheimer.' 
The wider release and demographic appeal of 'Barbie' further support the likelihood of it maintaining at least a 2x
lead in gross over 'Oppenheimer.' There is little evidence to suggest that 'Oppenheimer' will close the gap 
significantly enough to prevent 'Barbie' from doubling its gross.

5. Initial probability (prediction):

Given the strong evidence in favor of 'Barbie' doubling 'Oppenheimer's gross, the initial probability is high.

6. Evaluate whether your calculated probability is excessively confident or not confident enough:

The calculated probability seems confident, but it is based on concrete data and reports from reliable sources. 
However, it is important to consider the possibility of unforeseen factors that could affect the final outcome. 
While the probability is high, it should not be absolute to account for any potential uncertainties.

7. Output your final prediction:
*0.95*
\end{mdframed}
    \caption{A response from our fine-tuned model on the question \href{https://polymarket.com/event/will-barbie-gross-2x-more-than-oppenheimer-on-opening-weekend}{``Will `Barbie' gross 2x more than `Oppenheimer' on opening weekend?''}. The question resolved NO on July 24, 2023.  On July 20, 2023, the community gave 73\% and the response above gives $95\%$. In the original prompt (not shown here), our news retrieval provides projections that Barbie will likely outperform Oppenheimer at the box office. However, the model hallucinates them as facts (``[...] with 
figures exceeding the 2x mark''), resulting in overconfidence.}
    \label{fig:example-2}
\end{figure}

%% file: main.bbl
\begin{thebibliography}{}

\bibitem[Abolghasemi et~al., 2023]{abolghasemi2023humans}
Abolghasemi, M., Ganbold, O., and Rotaru, K. (2023).
\newblock Humans vs large language models: Judgmental forecasting in an era of advanced {AI}.
\newblock {\em arXiv preprint arXiv:2312.06941}.

\bibitem[Adam, 2020]{adam2020special}
Adam, D. (2020).
\newblock Special report: The simulations driving the world's response to {COVID}-19.
\newblock {\em Nature}, 580(7802):316--319.

\bibitem[Anthropic, 2023]{Claude}
Anthropic (2023).
\newblock Model card and evaluations for {Claude} models.
\newblock \url{https://www-cdn.anthropic.com/files/4zrzovbb/website/5c49cc247484cecf107c699baf29250302e5da70.pdf}.

\bibitem[Armstrong, 2001]{armstrong2001principles}
Armstrong, J.~S. (2001).
\newblock {\em Principles of Forecasting: a Handbook for Researchers and Practitioners}.
\newblock Springer.

\bibitem[Brier, 1950]{brier1950verification}
Brier, G.~W. (1950).
\newblock Verification of forecasts expressed in terms of probability.
\newblock {\em Monthly Weather Review}, 78(1):1--3.

\bibitem[Brown et~al., 2020]{brown2020language}
Brown, T., Mann, B., Ryder, N., Subbiah, M., Kaplan, J.~D., Dhariwal, P., Neelakantan, A., Shyam, P., Sastry, G., Askell, A., Agarwal, S., Herbert-Voss, A., Krueger, G., Henighan, T., Child, R., Ramesh, A., Ziegler, D., Wu, J., Winter, C., Hesse, C., Chen, M., Sigler, E., Litwin, M., Gray, S., Chess, B., Clark, J., Berner, C., McCandlish, S., Radford, A., Sutskever, I., and Amodei, D. (2020).
\newblock Language models are few-shot learners.
\newblock In {\em Advances in Neural Information Processing Systems (NeurIPS)}.

\bibitem[Chen et~al., 2023]{chen2023universal}
Chen, X., Aksitov, R., Alon, U., Ren, J., Xiao, K., Yin, P., Prakash, S., Sutton, C., Wang, X., and Zhou, D. (2023).
\newblock Universal self-consistency for large language model generation.
\newblock {\em arXiv preprint arXiv:2311.17311}.

\bibitem[Das et~al., 2024]{das2023decoder}
Das, A., Kong, W., Sen, R., and Zhou, Y. (2024).
\newblock A decoder-only foundation model for time-series forecasting.
\newblock {\em arXiv preprint arXiv:2310.10688}.

\bibitem[Deng et~al., 2023]{deng2023rephrase}
Deng, Y., Zhang, W., Chen, Z., and Gu, Q. (2023).
\newblock Rephrase and respond: Let large language models ask better questions for themselves.
\newblock {\em arXiv preprint arXiv:2311.04205}.

\bibitem[Dooley et~al., 2023]{dooley2023forecastpfn}
Dooley, S., Khurana, G.~S., Mohapatra, C., Naidu, S.~V., and White, C. (2023).
\newblock {ForecastPFN}: Synthetically-trained zero-shot forecasting.
\newblock In {\em Advanced in Neural Information Processing Systems (NeurIPS)}.

\bibitem[Gawlikowski et~al., 2021]{gawlikowski2021survey}
Gawlikowski, J., Tassi, C. R.~N., Ali, M., Lee, J., Humt, M., Feng, J., Kruspe, A., Triebel, R., Jung, P., Roscher, R., Shahzad, M., Yang, W., Bamler, R., and Zhu, X.~X. (2021).
\newblock A survey of uncertainty in deep neural networks.
\newblock {\em arXiv preprint arXiv:2107.03342}.

\bibitem[{Gemini Team}, 2023]{team2023gemini}
{Gemini Team} (2023).
\newblock Gemini: a family of highly capable multimodal models.
\newblock {\em arXiv preprint arXiv:2312.11805}.

\bibitem[Gneiting and Raftery, 2007]{gneiting2007strictly}
Gneiting, T. and Raftery, A.~E. (2007).
\newblock Strictly proper scoring rules, prediction, and estimation.
\newblock {\em Journal of the American Statistical Association}, 102(477):359--378.

\bibitem[Gruver et~al., 2023]{gruver2023large}
Gruver, N., Finzi, M.~A., Qiu, S., and Wilson, A.~G. (2023).
\newblock Large language models are zero-shot time series forecasters.
\newblock In {\em Advanced in Neural Information Processing Systems (NeurIPS)}.

\bibitem[Hanson, 2007]{hanson2007logarithmic}
Hanson, R. (2007).
\newblock Logarithmic markets coring rules for modular combinatorial information aggregation.
\newblock {\em The Journal of Prediction Markets}, 1(1):3--15.

\bibitem[Hendrycks et~al., 2021]{hendrycks2021unsolved}
Hendrycks, D., Carlini, N., Schulman, J., and Steinhardt, J. (2021).
\newblock Unsolved problems in {ML} safety.
\newblock {\em arXiv preprint arXiv:2109.13916}.

\bibitem[Izacard and Grave, 2021]{izacard2021leveraging}
Izacard, G. and Grave, {\'E}. (2021).
\newblock Leveraging passage retrieval with generative models for open domain question answering.
\newblock In {\em Proceedings of the 16th Conference of the European Chapter of the Association for Computational Linguistics (EACL)}.

\bibitem[Jiang et~al., 2024]{jiang2024mixtral}
Jiang, A.~Q., Sablayrolles, A., Roux, A., Mensch, A., Savary, B., Bamford, C., Chaplot, D.~S., de~las Casas, D., Hanna, E.~B., Bressand, F., Lengyel, G., Bour, G., Lample, G., Lavaud, L.~R., Saulnier, L., Lachaux, M.-A., Stock, P., Subramanian, S., Yang, S., Antoniak, S., Scao, T.~L., Gervet, T., Lavril, T., Wang, T., Lacroix, T., and Sayed, W.~E. (2024).
\newblock Mixtral of experts.
\newblock {\em arXiv preprint arXiv:2401.04088}.

\bibitem[Jin et~al., 2024]{jin2023time}
Jin, M., Wang, S., Ma, L., Chu, Z., Zhang, J.~Y., Shi, X., Chen, P.-Y., Liang, Y., Li, Y.-F., Pan, S., and Wen, Q. (2024).
\newblock {Time-LLM}: Time series forecasting by reprogramming large language models.
\newblock In {\em International Conference on Learning Representations (ICLR)}.

\bibitem[Jin et~al., 2021]{jin2021forecastqa}
Jin, W., Khanna, R., Kim, S., Lee, D.-H., Morstatter, F., Galstyan, A., and Ren, X. (2021).
\newblock {ForecastQA}: A question answering challenge for event forecasting with temporal text data.
\newblock In {\em Proceedings of the 59th Annual Meeting of the Association for Computational Linguistics and the 11th International Joint Conference on Natural Language Processing (ACL)}.

\bibitem[Lewis et~al., 2020]{lewis2020retrieval}
Lewis, P. S.~H., Perez, E., Piktus, A., Petroni, F., Karpukhin, V., Goyal, N., K{\"{u}}ttler, H., Lewis, M., Yih, W., Rockt{\"{a}}schel, T., Riedel, S., and Kiela, D. (2020).
\newblock Retrieval-augmented generation for knowledge-intensive {NLP} tasks.
\newblock In {\em Advances in Neural Information Processing Systems (NeurIPS)}.

\bibitem[Manifold, 2022]{Notion}
Manifold (2022).
\newblock Maniswap.
\newblock \url{https://manifoldmarkets.notion.site/manifoldmarkets/Maniswap-ce406e1e897d417cbd491071ea8a0c39}.

\bibitem[Metaculus, 2023]{Metaculus_2023}
Metaculus (2023).
\newblock Wisdom of the crowd vs. the best of the best of the best.
\newblock \url{https://www.metaculus.com/notebooks/15760/wisdom-of-the-crowd-vs-the-best-of-the-best-of-the-best}.

\bibitem[Min et~al., 2019]{min2019multi}
Min, S., Zhong, V., Zettlemoyer, L., and Hajishirzi, H. (2019).
\newblock Multi-hop reading comprehension through question decomposition and rescoring.
\newblock In {\em Proceedings of the 57th Annual Meeting of the Association for Computational Linguistics (ACL)}.

\bibitem[Nakano et~al., 2021]{nakano2021webgpt}
Nakano, R., Hilton, J., Balaji, S., Wu, J., Ouyang, L., Kim, C., Hesse, C., Jain, S., Kosaraju, V., Saunders, W., Jiang, X., Cobbe, K., Eloundou, T., Krueger, G., Button, K., Knight, M., Chess, B., and Schulman, J. (2021).
\newblock Web{GPT}: Browser-assisted question-answering with human feedback.
\newblock {\em arXiv preprint arXiv:2112.09332}.

\bibitem[Nie et~al., 2023]{nie2022time}
Nie, Y., Nguyen, N.~H., Sinthong, P., and Kalagnanam, J. (2023).
\newblock A time series is worth 64 words: Long-term forecasting with transformers.
\newblock In {\em International Conference on Learning Representations (ICLR)}.

\bibitem[Nye et~al., 2021]{nye2021show}
Nye, M., Andreassen, A.~J., Gur-Ari, G., Michalewski, H., Austin, J., Bieber, D., Dohan, D., Lewkowycz, A., Bosma, M., Luan, D., Sutton, C., and Odena, A. (2021).
\newblock Show your work: Scratchpads for intermediate computation with language models.
\newblock {\em arXiv preprint arXiv:2112.00114}.

\bibitem[OpenAI, 2023]{achiam2023gpt}
OpenAI (2023).
\newblock {GPT}-4 technical report.
\newblock {\em arXiv preprint arXiv:2303.08774}.

\bibitem[Polymarket, 2023]{Polymarket}
Polymarket (2023).
\newblock Polymarket/poly-market-maker: {Market Maker Keeper} for the polymarket {CLOB}.
\newblock \url{https://github.com/Polymarket/poly-market-maker}.

\bibitem[Rasul et~al., 2023]{rasul2023lag}
Rasul, K., Ashok, A., Williams, A.~R., Khorasani, A., Adamopoulos, G., Bhagwatkar, R., Biloš, M., Ghonia, H., Hassen, N.~V., Schneider, A., Garg, S., Drouin, A., Chapados, N., Nevmyvaka, Y., and Rish, I. (2023).
\newblock {Lag-Llama}: Towards foundation models for time series forecasting.
\newblock {\em arXiv preprint arXiv:2310.08278}.

\bibitem[Schoenegger and Park, 2023]{schoenegger2023large}
Schoenegger, P. and Park, P.~S. (2023).
\newblock Large language model prediction capabilities: Evidence from a real-world forecasting tournament.
\newblock {\em arXiv preprint arXiv:2310.13014}.

\bibitem[Shuster et~al., 2021]{shuster2021retrieval}
Shuster, K., Poff, S., Chen, M., Kiela, D., and Weston, J. (2021).
\newblock Retrieval augmentation reduces hallucination in conversation.
\newblock In {\em Findings of the Association for Computational Linguistics (Findings of EMNLP)}.

\bibitem[Tetlock and Gardner, 2015]{tetlock2015superforecasting}
Tetlock, P.~E. and Gardner, D. (2015).
\newblock {\em Superforecasting: The Art and Science of Prediction}.
\newblock Crown.

\bibitem[Touvron et~al., 2023]{touvron2023Llama}
Touvron, H., Martin, L., Stone, K., Albert, P., Almahairi, A., Babaei, Y., Bashlykov, N., Batra, S., Bhargava, P., Bhosale, S., Bikel, D., Blecher, L., Ferrer, C.~C., Chen, M., Cucurull, G., Esiobu, D., Fernandes, J., Fu, J., Fu, W., Fuller, B., Gao, C., Goswami, V., Goyal, N., Hartshorn, A., Hosseini, S., Hou, R., Inan, H., Kardas, M., Kerkez, V., Khabsa, M., Kloumann, I., Korenev, A., Koura, P.~S., Lachaux, M.-A., Lavril, T., Lee, J., Liskovich, D., Lu, Y., Mao, Y., Martinet, X., Mihaylov, T., Mishra, P., Molybog, I., Nie, Y., Poulton, A., Reizenstein, J., Rungta, R., Saladi, K., Schelten, A., Silva, R., Smith, E.~M., Subramanian, R., Tan, X.~E., Tang, B., Taylor, R., Williams, A., Kuan, J.~X., Xu, P., Yan, Z., Zarov, I., Zhang, Y., Fan, A., Kambadur, M., Narang, S., Rodriguez, A., Stojnic, R., Edunov, S., and Scialom, T. (2023).
\newblock Llama 2: Open foundation and fine-tuned chat models.
\newblock {\em arXiv preprint arXiv:2307.09288}.

\bibitem[Wang, 2023]{wang2023calibration}
Wang, C. (2023).
\newblock Calibration in deep learning: A survey of the state-of-the-art.
\newblock {\em arXiv preprint arXiv:2308.01222}.

\bibitem[Woo et~al., 2024]{woo2024unified}
Woo, G., Liu, C., Kumar, A., Xiong, C., Savarese, S., and Sahoo, D. (2024).
\newblock Unified training of universal time series forecasting transformers.
\newblock {\em arXiv preprint arXiv:2402.02592}.

\bibitem[Yan et~al., 2024]{yan2023autocast}
Yan, Q., Seraj, R., He, J., Meng, L., and Sylvain, T. (2024).
\newblock Autocast++: Enhancing world event prediction with zero-shot ranking-based context retrieval.
\newblock In {\em International Conference on Learning Representations (ICLR)}.

\bibitem[Zhang et~al., 2018]{zhang2018formal}
Zhang, Y., Chen, X., and Park, D. (2018).
\newblock Formal specification of constant product (xy= k) market maker model and implementation.
\newblock {\em White paper}.

\bibitem[Zhu et~al., 2024]{zhu2023large}
Zhu, Y., Yuan, H., Wang, S., Liu, J., Liu, W., Deng, C., Dou, Z., and Wen, J.-R. (2024).
\newblock Large language models for information retrieval: A survey.
\newblock {\em arXiv preprint arXiv:2308.07107}.

\bibitem[Zou et~al., 2022]{zou2022forecasting}
Zou, A., Xiao, T., Jia, R., Kwon, J., Mazeika, M., Li, R., Song, D., Steinhardt, J., Evans, O., and Hendrycks, D. (2022).
\newblock Forecasting future world events with neural networks.
\newblock In {\em Advances in Neural Information Processing Systems (NeurIPS)}.

\end{thebibliography}
